\documentclass{egpubl}
\usepackage{eurovis2021}

\SpecialIssuePaper         % uncomment for final version of Computer Graphics Forum, special issue
\usepackage[T1]{fontenc}
\usepackage{dfadobe}  

\usepackage{cite}  % comment out for biblatex with backend=biber
\BibtexOrBiblatex
\electronicVersion
\PrintedOrElectronic
\ifpdf \usepackage[pdftex]{graphicx} \pdfcompresslevel=9
\else \usepackage[dvips]{graphicx} \fi

\usepackage{egweblnk}
\usepackage{subfig}
\usepackage{amsmath}
\usepackage{amssymb}

\usepackage{color}
\definecolor{edit}{RGB}{23,55,180}

\title[Compressive Neural Representations]%
      {Compressive Neural Representations of Volumetric Scalar Fields}

\author[Y. Lu \& K. Jiang \& J. A. Levine \& M. Berger]
{\parbox{\textwidth}{\centering Y. Lu$^{1}$, K. Jiang$^{2}$, J. A. Levine$^{2}$\orcid{0000-0002-4302-1704}, and M. Berger$^{1}$
        }
        \\
{\parbox{\textwidth}{\centering $^1$Department of Electrical Engineering and Computer Science, Vanderbilt University,  USA\\
         $^2$Department of Computer Science, University of Arizona, USA
       }
}
}
\begin{document}

% uncomment for using teaser
\teaser{
 \includegraphics[width=\linewidth]{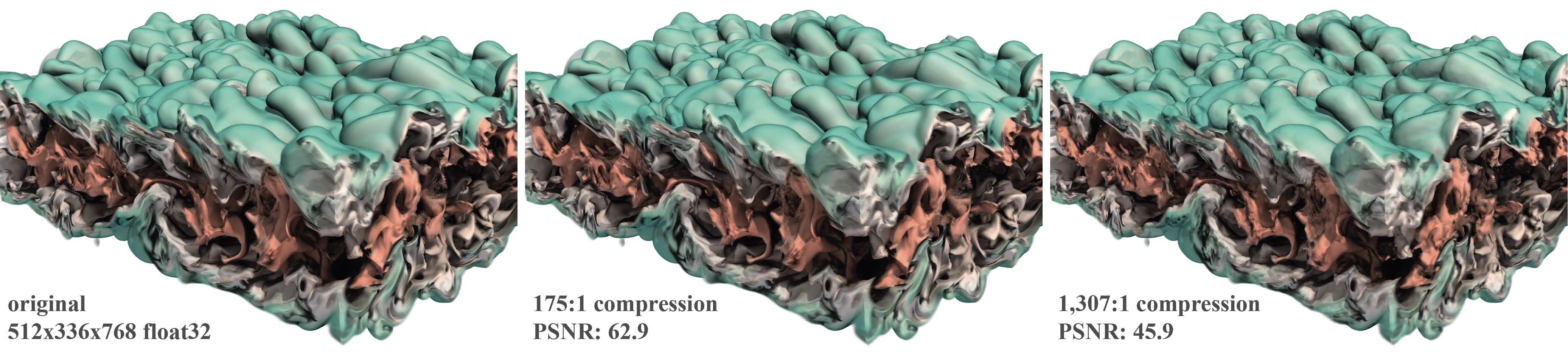}
 \centering
 \caption{Our approach represents a scalar field as a neural network that conditions on a point in the field's domain and produces a scalar value. We obtain highly compressive volume representations in this manner, here showing two levels of compression (middle, right) for the \textsf{jet} volume (left), where even for extreme compression ratios (right, $1307:1$), predominant features of the volume are preserved.}
 \label{fig:teaser}
}

\maketitle
%-------------------------------------------------------------------------
\begin{abstract}
We present an approach for compressing volumetric scalar fields using implicit neural representations. Our approach represents a scalar field as a learned function, wherein a neural network maps a point in the domain to an output scalar value. By setting the number of weights of the neural network to be smaller than the input size, we achieve compressed representations of scalar fields, thus framing compression as a type of function approximation. Combined with carefully quantizing network weights, we show that this approach yields highly compact representations that outperform state-of-the-art volume compression approaches. The conceptual simplicity of our approach enables a number of benefits, such as support for time-varying scalar fields, optimizing to preserve spatial gradients, and random-access field evaluation. We study the impact of network design choices on compression performance, highlighting how simple network architectures are effective for a broad range of volumes.
%-------------------------------------------------------------------------
%  ACM CCS 1998
%  (see https://www.acm.org/publications/computing-classification-system/1998)
% \begin{classification} % according to https://www.acm.org/publications/computing-classification-system/1998
% \CCScat{Computer Graphics}{I.3.3}{Picture/Image Generation}{Line and curve generation}
% \end{classification}
%-------------------------------------------------------------------------
%  ACM CCS 2012
%  (see https://www.acm.org/publications/class-2012)
%The tool at \url{http://dl.acm.org/ccs.cfm} can be used to generate
% CCS codes.
%Example:


\end{abstract}

%-------------------------------------------------------------------------
\section{Introduction}

% what is the problem, why is it important
The visualization of large-scale field-based data is a fundamental component to many post hoc analyses. Field data, in particular scalar fields, often arise from numerical simulations of scientific phenomena, which require high (3D) spatial and temporal resolution to accurately resolve domain-specific features of interest. The size of such large-scale data presents a number of challenges for visualization, ranging from bandwidth constraints, disk storage, data accessibility, and interactivity. Compression of volumetric scalar fields remains an important tool that can help mitigate these challenges.  For visualization purposes, the data is often so large that \emph{lossy} compression methods are required to enable analysis. The main purpose of lossy methods is to obtain compact volumetric representations in which features of the data remain visually perceptible, at the cost of sacrificing some data precision.

% traditional approaches to lossy volumetric compression: transforms that yield compressive representations; limitation: each type of transform has certain assumptions that might be satisfied to varying degrees in practice
Within the visualization community, the lossy compression of volumetric scalar fields has largely been based on transforms that admit compressive representations,  where transform coefficients can be discarded and/or aggressively quantized with minimal loss in data fidelity. A commonality to these methods is the reliance on a rectilinear grid for which predefined bases may be constructed, e.g. Fourier~\cite{yeo1995volume} or Wavelet bases~\cite{woodring2011revisiting}, or where data-dependent bases may be derived~\cite{suter2013tamresh,ballester2016lossy}. However, the success of these methods heavily depends on whether a provided scalar field is a good match for the assumptions made by the transform, e.g. the field is largely composed of low-frequencies for Fourier bases, or admits a low-rank decomposition for factorization-based transforms~\cite{ballester2019tthresh}. In practice, these assumptions are satisfied to varying degrees, leading to variability in compression ratios and approximation quality amongst existing methods.

% our approach: function-space neural networks for approximating scalar fields; use of deep neural networks: no such assumptions on data characteristics, let the network learn what is necessary to approximate the given field
In this work we propose a learning-based method for compressing scalar fields. We leverage neural networks that are designed to map a continuously-defined position from the domain to a scalar value in the range~\cite{park2019deepsdf,mescheder2019occupancy}. The use of neural networks places little assumptions on data characteristics; instead, the network learns what is necessary to approximate the given field. In particular, by limiting the capacity of the network, such that the number of weights of the network is less than the volume resolution, we obtain compressive volume representations that are optimized to best approximate the field at its sampled values. Hence, the neural network \emph{is} the compressed volume representation -- the level of compression, in part, follows from the number of network weights, and the original sampled field can be reconstructed by evaluating the neural network at the given positions.

% what we study; simple network designs for compressive representations: fully-connected layers with periodic activation functions (SIREN) are sufficient, no normalization needed; residual connections key permit deeper, and better, approximations; SIREN weights tend to be normally-distributed, careful quantization of network weights leads to further compression with little hit on performance
Our neural network design is conceptually straightforward. We build on the approach of SIREN~\cite{sitzmann2020implicit}, wherein fully-connected layers and periodic activation functions (using sinusoids) comprise the network architecture.  
Periodic activation functions, coupled with careful initialization for stable optimization, have been recently shown to outperform more advanced architectures, e.g. frequency-based embeddings with ReLU activations~\cite{mildenhall2020nerf}, for a variety of representation tasks.  
We illustrate how residual connections that preserve SIREN activation distributions lead to networks that are robust across architecture choices. Further, we observe that the learned, per-layer, weights of SIREN are distributed in such a manner that they may be quantized to a small number of bits, with minimal impact on performance.

% key benefits: highly-compact representations of scalar fields, improve over STAR; permit optimization of higher-order information, e.g. spatial gradients -> reframe compression not just as approximating field values, but quantities that we care about for visual exploration, (gradients -> normals for isosurfacing); time-varying data, and spatiotemporal gradients
By limiting network capacity, and performing weight quantization, we obtain highly compact representations of scalar fields, as demonstrated in Fig.~\ref{fig:teaser}. Further, the conceptual simplicity of our approach provides a number of benefits. Due to the use of periodic activation functions, we can target the optimization of higher-order derivatives, as studied in Sitzmann et al.~\cite{sitzmann2020implicit}. In particular, when spatial gradients are available, we can use this information to regularize the network. This leads to better-behaved scalar fields whose isosurfaces are less noisy under high compression ratios. We also show how it is straightforward to adapt our approach to compressing time-varying scalar fields simply by adjusting the network inputs. Furthermore, the networks enable field access at arbitrary points in the domain, eschewing the reconstruction of the entire volume at once -- we show how this can benefit visualization applications such as volume rendering.

Summarizing, the main contributions of our approach are:
\begin{enumerate}
    \item A compression technique for volumetric scalar fields based on implicit neural representations.
    \item An evaluation of this compression approach, directly comparing it to a recent state-of-the-art technique (\textsf{tthresh}~\cite{ballester2019tthresh}) as well as exploring implementation design choices; and
    \item Experimental results on compressing volumes, including studying gradient preservation and preserving time-varying data.
\end{enumerate}

%-------------------------------------------------------------------------
\section{Related Work}

\subsection{Compression of Volumetric Data}

Lossy compression of large scale volumes has been an important topic in recent years, particularly as our ability to simulate and acquire data grows.  Many of the early works in this area focused on using discrete cosine transforms~\cite{yeo1995volume} or wavelet-based compression~\cite{guthe2002interactive,ihm1999wavelet,muraki1993volume,woodring2011revisiting}, often specifically focusing on interactive rendering applications~\cite{rodriguez2013survey,schneider2003compression}.  By using such transforms, these techniques enable compression by allowing the end user to separate features within a transformed space (i.e. removing high frequencies while preserving low frequencies) and ultimately sparsify the data representation.  ZFP performs transformations customized at the block level to achieve compression with fast I/O access~\cite{lindstrom2014fixed}.  More recent techniques, such as TAMRESH~\cite{suter2013tamresh} and TTHRESH~\cite{ballester2019tthresh} employ tensor decomposition~\cite{ballester2016lossy}.  TTHRESH is particularly notable as a recent state-of-the-art compression technique that we directly compare our results against in this work.

Other techniques focus on data fitting as the workhorse for compression.  For example, ISABELA~\cite{lakshminarasimhan2011compressing} and SZ~\cite{di2016fast,tao2017significantly} use curve fitting to approximate the input data.  Liang et al.~outperform SZ and ZFP using an improved Lorenzo prediction~\cite{liang2018error}.  Data fitting is related to our neural network based approach, although the method to achieve the fit is different.  Using data models can be a powerful mechanism for providing a compression technique. Peterka et al.~\cite{peterka2018foundations} study the use of nonuniform rational B-spline functions (NURBS) for compressive volume representations via adaptive refinement of control points. This work shows the benefit of smooth representations for analytically-defined gradients, however an important distinction in our approach is that we can directly optimize for gradients. Other examples focus on models that preserve specific aspects of the data, e.g. topological features~\cite{soler2018topologically}, graph-based models like SQ~\cite{iverson2012fast}, and dictionaries~\cite{diaz2020interactive}.  COVRA uses an octree decomposition to model data blocks which are compressed at multiple resolutions~\cite{gobbetti2012covra}.  Data fitting broadly considers the tradeoff between how many samples are used and how much precision is devoted to each sample.  Recently, Hoang et al.~provide a precision-resolution tree that is used for efficient selection of either fewer data samples or data samples encoded with fewer bits~\cite{hoang2020efficient}.

\begin{figure*}[!t]
    \centering
    \includegraphics[width=0.96\linewidth]{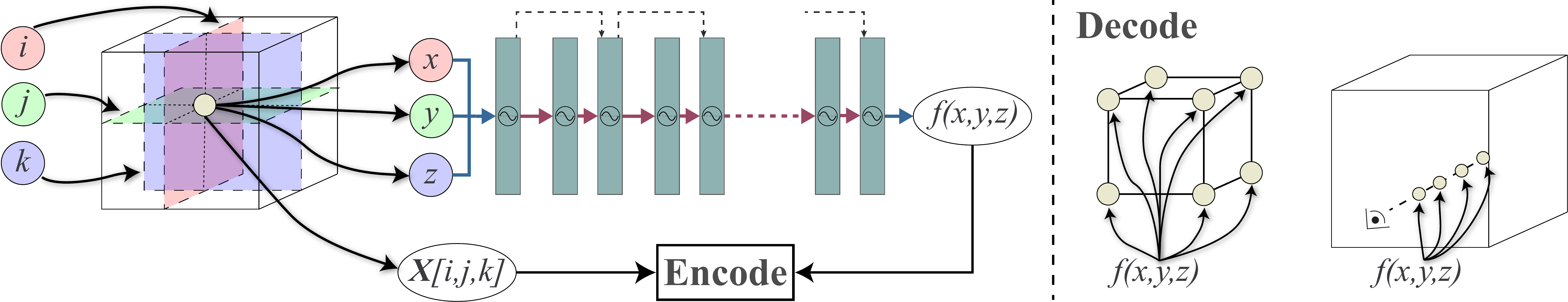}
    \caption{An overview of our compression method, split into the steps of encoding (left) and decoding (right). We encode a provided scalar field with a coordinate-based neural network, where the network $f$ is tasked with predicting the field's value $\mathbf{X}[i,j,k]$ given coordinates $(x,y,z)$. In the network diagram, red arrows represent fully-connected layers whose weights are quantized, blue arrows are unquantized fully-connected layers, and dashed black arrows represent residual connections. Decoding amounts to function evaluation, where we may reconstruct the field at grid vertices, or arbitrary positions, and thus perform volume rendering with the network in lieu of the sampled field.}
    \label{fig:overview}
\end{figure*}

\subsection{Deep Learning for Volume Representation}

% Talk about deep learning in visualization, e.g. insitunet
%Maybe FlowNet?
Recently, numerous researchers have deployed deep learning and used neural networks to enable a variety of visualization tasks.  Berger et al.~utilize generative adversarial networks (GANs) to analyzing the space of visual parameters in a volume renderer, providing a new interface for transfer function design~\cite{berger2018generative}.  He et al.~generalize the idea of learning conditioned on parameter spaces to include ensembles of volumes with InSituNet~\cite{he2019insitunet}.  Han et al.~encode multivariate volume data for variable-to-variable translation~\cite{han2020v2v}.  Jakob et al.~consider the problem of interpolation using neural networks defined on a large ensemble of vector fields~\cite{jakob2020fluid}.  While the applications are different, all of these works share a common thread in that they construct representations that blend visualization tasks (e.g. rendering, interpolation) with data representations.

Closely related to compression is the task of super-resolution, for which many techniques have been developed in the computer vision communities.  More recently, researchers have shown that deep learning can be used to achieve super-resolution on volumetric data.  Xie et al.~developed tempoGAN for super-resolution of volumetric fluids in computer graphics~\cite{xie2018tempogan}.  Weiss et al.~developed new techniques for isosurface rendering that similarly used deep learning for super-resolution~\cite{weiss2019volumetric}.  Han and Wang developed neural networks for super-resolution in temporal~\cite{han2019tsr} and spatial~\cite{han2020ssr} super-resolution, and Guo et al.~achieved spatial super-resolution in vector fields~\cite{guo2020ssr}.  Notably, many of these techniques are tested against possible compression techniques, but none of the above were specifically designed for compression.  Thus, while they are compressive, their main strengths lie elsewhere.

Recently, the machine learning community has explored using deep learning to construct implicit representations, which can be used to model a variety of data.  We utilize a similar approach in our network design, and as seen in recent works this can achieve superior performance.  Park et al.~developed DeepSDF for representing shape using signed distance fields~\cite{park2019deepsdf}.  Recent extensions to this work include overfit neural networks~\cite{davies2020overfit} and DualSDF~\cite{hao2020dualsdf} that allow for tradeoffs in the representation precision and flexbility.  Similarly, Chen and Zhang use implicit fields for generative shape modeling~\cite{chen2019learning}.  Alternatively, instead of training to produce a signed distance field, Mescheder et al. propose to construct occupancy fields~\cite{mescheder2019occupancy}.  Besides just representing shape, other recent applications include texture synthesis~\cite{oechsle2019texture} and view synthesis~\cite{mildenhall2020nerf}.  As this is an emerging area, we are still discovering new ways to improve these networks.  Notable recent works that informed our network design include using periodic activations~\cite{sitzmann2020implicit} and Fourier feature mappings~\cite{tancik2020fourier}. Our approach is distinct from prior works in that we investigate how to make implicit neural representations compressive for arbitrary scalar fields~\cite{davies2020overfit,sitzmann2020implicit,tancik2020fourier}.

%-------------------------------------------------------------------------
\section{Compression Method}

The basic idea behind our compression approach is to represent a provided scalar field as a learned function, one that is parameterized by a set of weights whose size is smaller than the field's size.  Fig.~\ref{fig:overview} shows an overview. Specifically, we assume the input to our method is a scalar field sampled on a regular grid, which we treat as a $d$-tensor $\mathbf{X} \in \mathbb{R}^{s_1 \times s_2 \times \cdots \times s_d}$. When $d=3$ this represents a scalar field whose domain is a 3D Cartesian coordinate system, while $d=4$ may represent a time-varying scalar field in 4 dimensions. We associate each sample of the field with an indexing tuple of integers $\mathbf{i} = (i_1,i_2,\cdots,i_d)$, an integer for each dimension, such that $\mathbf{X}[i_1,i_2,\cdots,i_d]$ returns the value of the field at this sample, and further, assume this index is associated with a real-valued $d$-dimensional point from the field's domain, $\mathbf{p}_{\mathbf{i}} \in \mathbb{R}^d$.

Our goal is to learn a function $f_{\Theta} : \mathbb{R}^d \rightarrow \mathbb{R}$, such that for a given sample $\mathbf{i} = (i_1,i_2,\cdots,i_d)$ and corresponding point $\mathbf{p}_{\mathbf{i}}$, we would like $f_{\Theta}(\mathbf{p}_{\mathbf{i}})$ to be as close to $\mathbf{X}[i_1,i_2,\cdots,i_d]$ as possible. The vector $\Theta \in \mathbb{R}^m$ denotes the function's set of parameters. Assuming its size $m$ is less than the size of the field, defined as $C = \prod_{j=1}^d s_j$, then we can obtain a representation of the field that (a) serves as a good approximation to $\mathbf{X}$, and (b) requires smaller storage.

What specific form should the function take? In order to handle complex volumetric fields typically produced in numerical simulations, we require functions that place as few assumptions as possible on the characteristics of the fields, e.g. smoothness, spectral properties, or sampling rate. To this end, we define the functions as deep neural networks, building off of recent work in implicit neural representations for shape modeling~\cite{park2019deepsdf,mescheder2019occupancy,sitzmann2020implicit}. Notably, these methods can model fine-grained details, e.g. intricate geometric structures of shapes, only limited by the capacity of the network rather than the resolution of some underlying regular grid, as is commonly used with convolutional networks~\cite{han2019tsr}. We argue that such coordinate-based neural networks are a good fit for arbitrary volumetric fields, not just shape-based data.  By framing compression as optimization, we allow the network to learn what is necessary for approximating the sampled field under a compression budget. In this paper we study how to effectively construct such functions, considering the trade-offs in approximation quality and level of compression. At a high-level, our approach follows a standard compression setup, comprised of 2 steps: an \emph{encode} step to find the function, and a \emph{decode} step that allows us to reconstruct the sampled field.

\subsection{Neural Encoder}

The purpose of the encode step is to train the neural network to accurately model the field at its provided samples. More specifically, we aim to minimize the following mean-squared error loss:
\begin{equation}
    \min_{\Theta} \sum_{\mathbf{i}} \left( f_{\Theta}(\mathbf{p}_{\mathbf{i}}) - \mathbf{X}[i_1,i_2,\cdots,i_d] \right)^2.
    \label{eq:loss}
\end{equation}
In practice, the loss function is optimized by performing stochastic gradient descent, assembling minibatches at each iteration of training by randomly selecting samples from the provided field.

\begin{figure}[!t]
    \centering
    \subfloat[\label{subfig:layer_asteroid}\textsf{Asteroid}]{
         \includegraphics[width=0.47\linewidth]{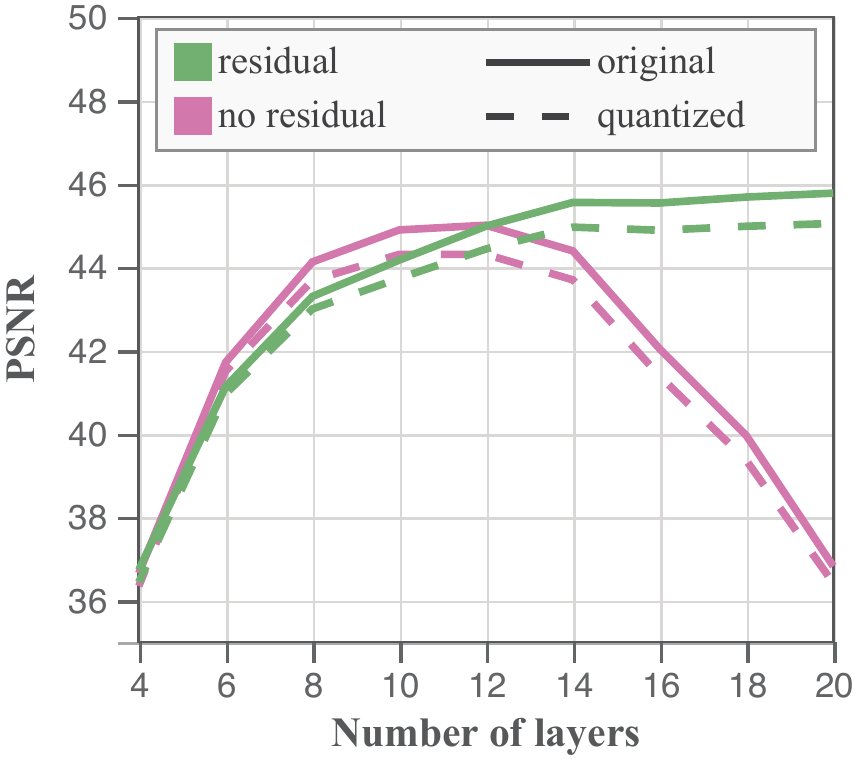}
	}
    \subfloat[\label{subfig:layer_ionization}\textsf{Ionization}]{
         \includegraphics[width=0.47\linewidth]{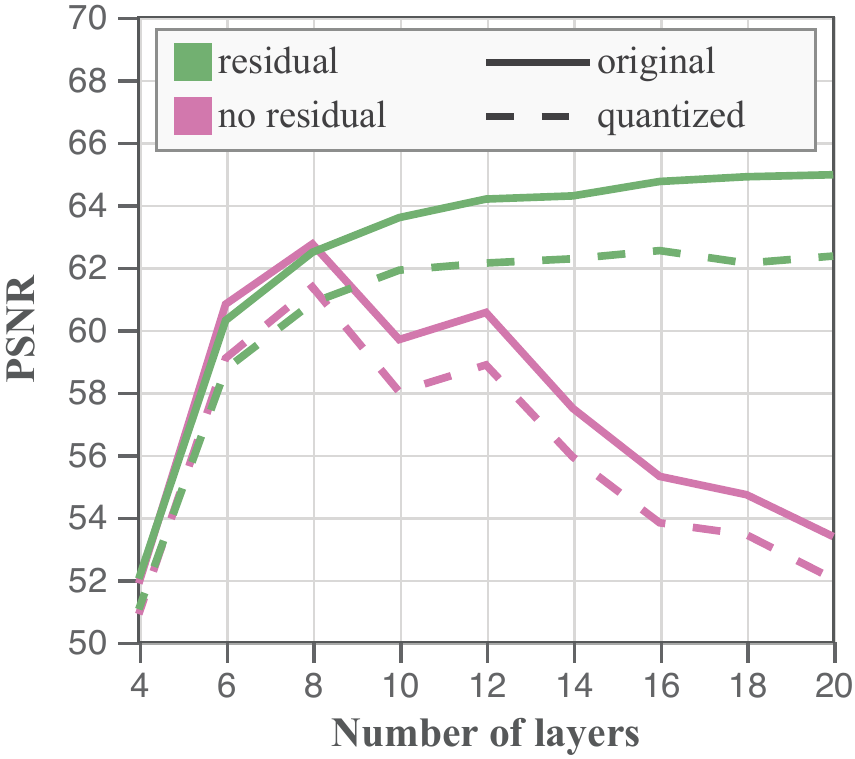}
	}
    \caption{We show the impact of residual connections (green) vs. non-residual connections (red) as a function of network depth, as well as the impact of network weight quantization (dashed lines), fixing the number of network weights $m$ to be $\frac{C}{50}$. We find that residual connections are more robust to depth and improve performance, while weight quantization leads to a small drop in performance, but is volume-dependent.}
    \label{fig:layers}
\end{figure}

What remains is the design of the network. We opt for simplicity in network architectures, to avoid the brittleness of hyperparameter tuning and support a diverse set of volumetric fields. To this end, we build off of the approach of SIREN~\cite{sitzmann2020implicit}, where a neural network is defined by a sequence of fully-connected layers that use sinusoidal activation functions -- no normalization, e.g. batch normalization~\cite{ioffe2015batch}, is necessary for successful training. SIREN has several benefits. First, as demonstrated by Sitzmann et al., training is ensured stable through a principled initialization scheme that controls for the distribution of activations at layers, namely that all layers are arcsine-distributed. Secondly, the use of sinusoidal activations admits functions that are $C^{\infty}$ differentiable. Our approach takes advantage of both of these properties.

Given the network design of SIREN, to setup the network we need only define the fully-connected layers. To this end, our method accepts two inputs: the number of layers $l$, and the total number of network weights $m$. We derive an integer $k$ such that the first weight matrix has size $k \times d$, the last weight matrix has size $d \times 1$, and all intermediary weight matrices are of size $k \times k$, such that the total number of weights of the network is approximately $m$. Bias vectors are similarly derived from inputs $m$ and $l$. All weights are of 32-bit floating-point precision. The resulting set of weight matrices and bias vectors, collectively, represents our vector of parameters $\Theta$, e.g. the compressed representation. Note that it is also possible to vary the sizes of the weight matrices, e.g. to gradually increase the number of hidden units as a function of layer depth. Experimentally, we found this made little difference, thus for simplicity we chose a fixed number of hidden units in intermediary layers.

The number of layers, $l$, should be large enough to handle the modeling of arbitrarily complex fields. In practice, however, setting $l$ too large can be detrimental, posing challenges for training due to vanishing gradients. As an example, we show results for two different volumes in Fig.~\ref{fig:layers} (red solid lines), varying the number of layers, where the quality of the volume is measured through peak signal-to-noise ratio (PSNR), the mean-squared error normalized by the data range. Note that, at a certain point, as the model depth increases performance decreases. Further, the network depth that leads to highest PSNR varies between the two volumes.

\begin{figure}[!t]
    \centering
    \subfloat[\label{subfig:weights_3} Layer 3]{
         \includegraphics[width=0.47\linewidth]{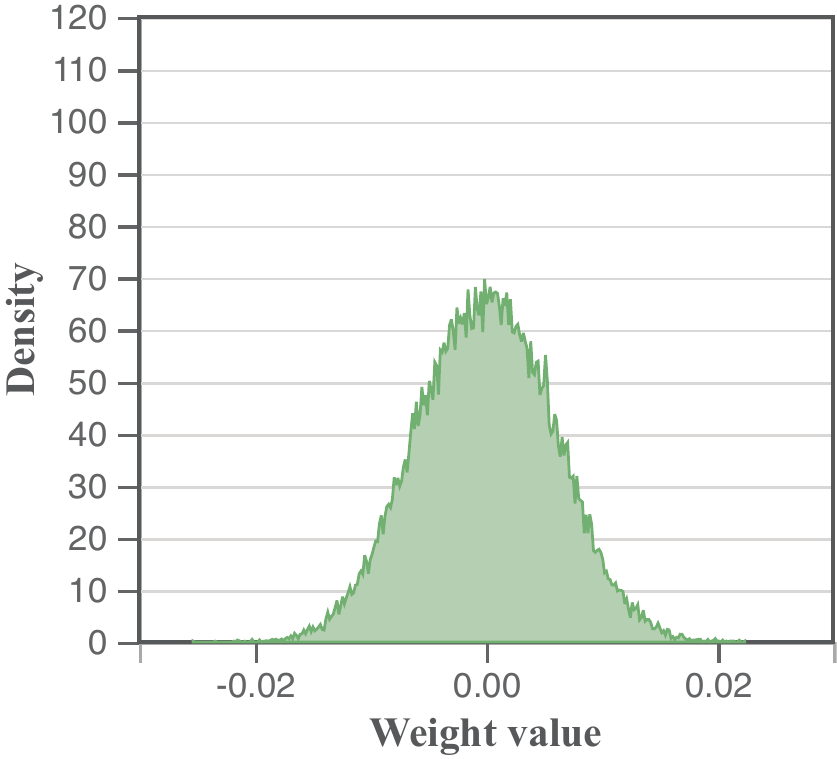}
	}
    \subfloat[\label{subfig:weights_9} Layer 9]{
         \includegraphics[width=0.47\linewidth]{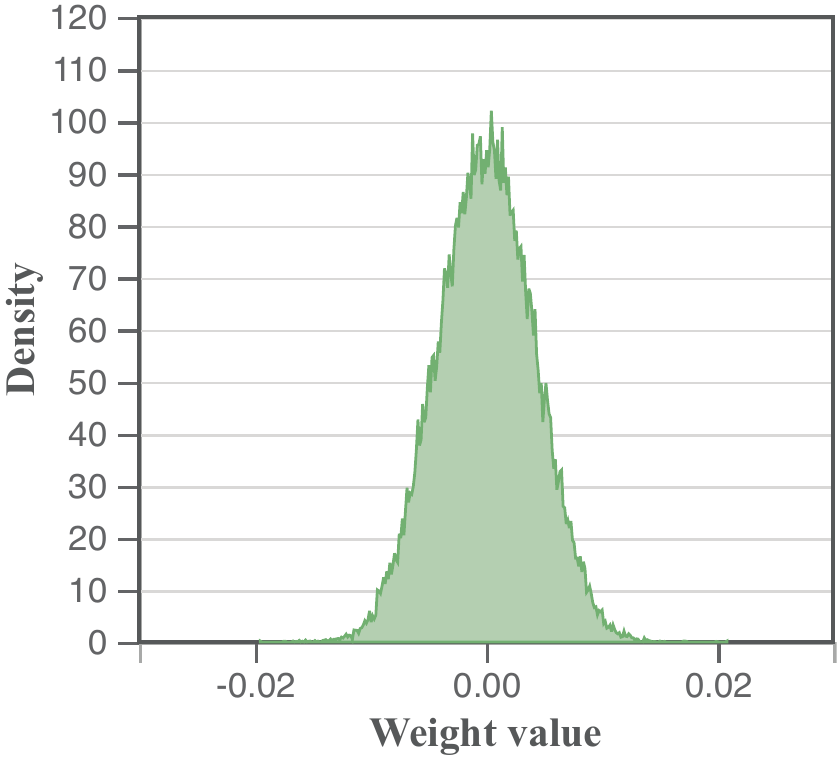}
	}
	\newline
    \subfloat[\label{subfig:weights_15} Layer 15]{
         \includegraphics[width=0.47\linewidth]{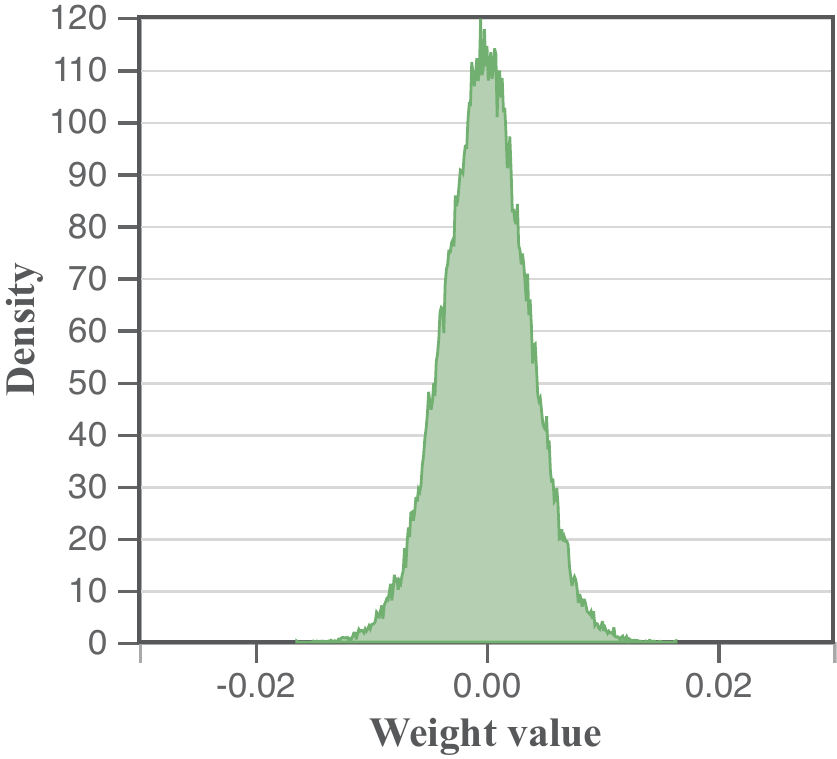}
	}
    \subfloat[\label{subfig:ionization_bits} Quantization v. Depth]{
         \includegraphics[width=0.47\linewidth]{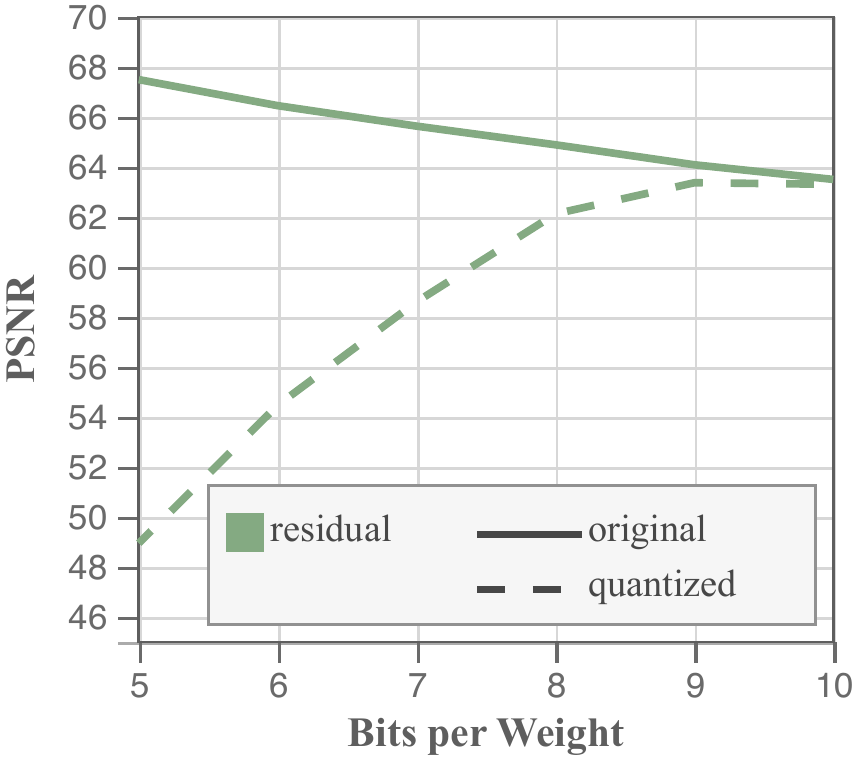}
	}
    \caption{In (a-c) we show the distribution of weights for our network trained on the \textsf{Ionization} volume for layers 3, 9, and 15, number of weights $m = \frac{C}{50}$. We find that the distribution is approximately zero-mean normal, with slight variations, hence amenable to quantization. In (d) we vary the number of bits used for quantization, adjusting the number of weights $m$ to ensure an approximately equal compression ratio. We find that $9$ bits is a good trade-off between number of network weights, and level of quantization.}
    \label{fig:quantization}
\end{figure}

In order to ensure our models are robust with respect to the number of layers across different volumes, we enrich SIREN with residual connections~\cite{he2016identity}. Specifically, we modify the identity mappings of He et al.~\cite{he2016identity} to ensure that the distribution of activations in each layer are within the range $[-1,1]$, in accordance with the assumptions of SIREN~\cite{sitzmann2020implicit}. Each residual block in our network takes the following form, omitting bias vectors for clarity (c.f. Fig.~\ref{fig:overview} dashed black arrows):
\begin{equation}
\mathbf{a}_{i+1} = \frac{1}{2}\left(\mathbf{a}_i + \sin\left(\mathbf{M}^2_{i+1} \sin(\mathbf{M}^1_{i+1} \mathbf{a}_i) \right) \right),
\end{equation}
where $\mathbf{a}_i$ represents a vector of activations from a previous block $i$, $\mathbf{a}_{i+1}$ is a vector of activations at the next block $i+1$, and $\mathbf{M}^1_{i+1}$ and $\mathbf{M}^2_{i+1}$ are a pair of learnable matrices associated with block $i+1$. In Fig.~\ref{fig:layers} (green solid lines) we show the effect of residual connections -- note we treat each block as containing 2 layers. Note that (a) training is more stable as the network depth increases, and (b) we obtain a boost in performance over non-residual connections. Consequently, unless otherwise stated, all of our networks employ 8 residual blocks.

\subsubsection{Network Weight Quantization}

Thus far, the compression ratio of our method can be expressed as $\frac{C}{m}$, assuming the network weights and volume are of the same precision. However, further compression can be gained through quantizing the network weights. Weight quantization has been considered for classification-based convolutional networks~\cite{han2015deep}, but to the best of our knowledge, not yet studied for coordinate-based MLPs. For our SIREN-based network, we make several observations:
\begin{enumerate}
    \item The first and last layers of the network are parameterized by small weight matrices, as discussed above. Experimentally, we found that they require high precision for good performance and thus we do not quantize the matrices (c.f. Fig.~\ref{fig:overview} blue arrows).
    \item Instead, most of the parameters of our network exist in intermediary layers, parameterized by matrices of $k \times k$. It is these matrices that we quantize (c.f. Fig.~\ref{fig:overview} red arrows).
    \item Since the distribution of activations in SIREN take a specific form, we find that the distribution of weights in each layer take on, approximately, a normal distribution, with some small variations. Fig.~\ref{fig:quantization}(a)-(c) demonstrates this for three different layers, where empirically, we find that deeper layers have lower spread, and a higher concentration of values around zero.
\end{enumerate}
Motivated by these observations, we employ the technique of Han et al.~\cite{han2015deep} and cluster the weights, individually, for each intermediary layer using k-means. We set the number of clusters to be the precision at which we would like to represent the weights, namely if we wish to represent the weights with a precision of $b$ bits, then the number of clusters will be set to $2^b$. After clustering, we assign each weight its corresponding $b$-bit index, and store these, alongside the corresponding floating point-precision cluster centers, as the quantized weight representation.

What should the number of bits, $b$, be set to in practice? In Fig.~\ref{fig:layers} we show the two network options with their corresponding quantized representations (as dashed lines), setting $b$ to 8. We find that the drop in performance in quantization is dependent on the specific volume, with the \textsf{ionization} volume most impacted. Thus, we cannot be too aggressive in quantization, yet there exists a size trade-off in the precision at which to represent weights, and the number of weights to assign to the network. Based on this, in Fig.~\ref{fig:quantization}(d) we plot the PSNR for a sequence of networks that all have approximately the same size, but consider both the number of weights and weight quantization. We increase the number of bits used for quantization along the x-axis, and compensate by decreasing the number of weights of the network. We observe that at $9$ bits, e.g. $512$ clusters, the network takes a minimal drop in PSNR -- similar results are observed in different volumes. Consequently, all networks in the paper represent weights of intermediary layers with 9 bits.

\subsubsection{Gradient Regularization}
\label{subsubsec:gradreg}

As previously mentioned, the use of sinusoidal activation functions in our network implies that we can take arbitrary-order derivatives of the function. Thus we can optimize for the derivatives of the function, in addition to just the provided field values. This allows us to obtain compressed fields whose higher-order information is preserved. Specifically, we modify our loss in Eq.~\ref{eq:loss} to ensure that the spatial gradient of the function is close to provided gradients of the original scalar field:
\begin{equation}
\begin{aligned}
\min_{\Theta} \ \ \ & \sum_{\mathbf{i}} \left( f_{\Theta}(\mathbf{p}_{\mathbf{i}}) - \mathbf{X}[i_1,i_2,\cdots,i_d] \right)^2 + \\ 
 & \lambda \lVert \nabla f_{\Theta}(\mathbf{p}_{\mathbf{i}}) - \mathbf{X}^{'}[i_1,i_2,\cdots,i_d] \rVert^2_ 2,
\end{aligned}
\label{eq:grad-loss}
\end{equation}
where $\mathbf{X}^{'}$ denotes the (provided) scalar field gradient, and we set $\lambda = 0.05$, which we found to give a good balance between the gradient and scalar value fit. In some cases, $\mathbf{X}^{'}$ may be accompanied by the field, e.g. within FEM numerical simulations gradients are often computed from the finite element basis. In other cases, we may numerically approximate gradients using finite differencing schemes, however for certain volumes, e.g. acquired medical images, the estimated gradients may not be reliable for regularization.

\begin{table}[!t]
\caption{\label{tab:grad}We compare the impact of gradient regularization (``grad'') without using such regularization (``no grad'') for the \textsf{jet} volume, for a compression ratio of $677:1$. We find this regularization yields gradients closer to the target (Grad PSNR) with minimal drop in field approximation (PSNR).}
\centering
\begin{tabular}{|r|c|c|c|}
\hline
\textbf{Loss}    & \textbf{PSNR} & \textbf{FD-Grad PSNR} & \textbf{Net-Grad PSNR} \\ \hline
no grad & 51.8 & 52.0         & 50.3          \\ \hline
grad    & 51.6 & 54.7         & 55.5          \\ \hline
\end{tabular}
\end{table}

\begin{figure}[!t]
    \centering
    \subfloat[\label{subfig:jet_iso_gt}ground truth]{
         \includegraphics[width=0.31\linewidth]{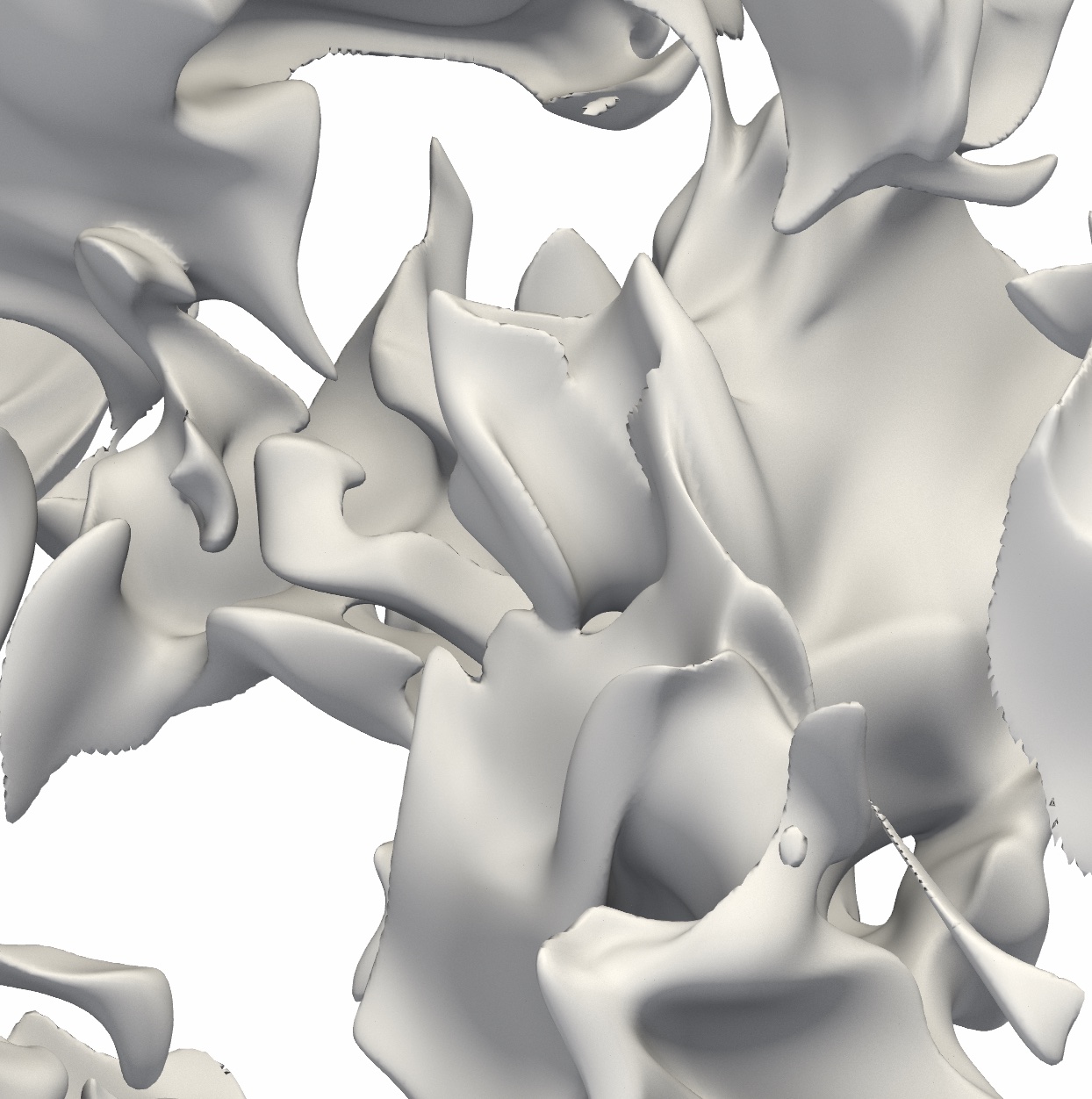}
	}
    \subfloat[\label{subfig:jet_iso_nograd}no gradient reg.]{
         \includegraphics[width=0.31\linewidth]{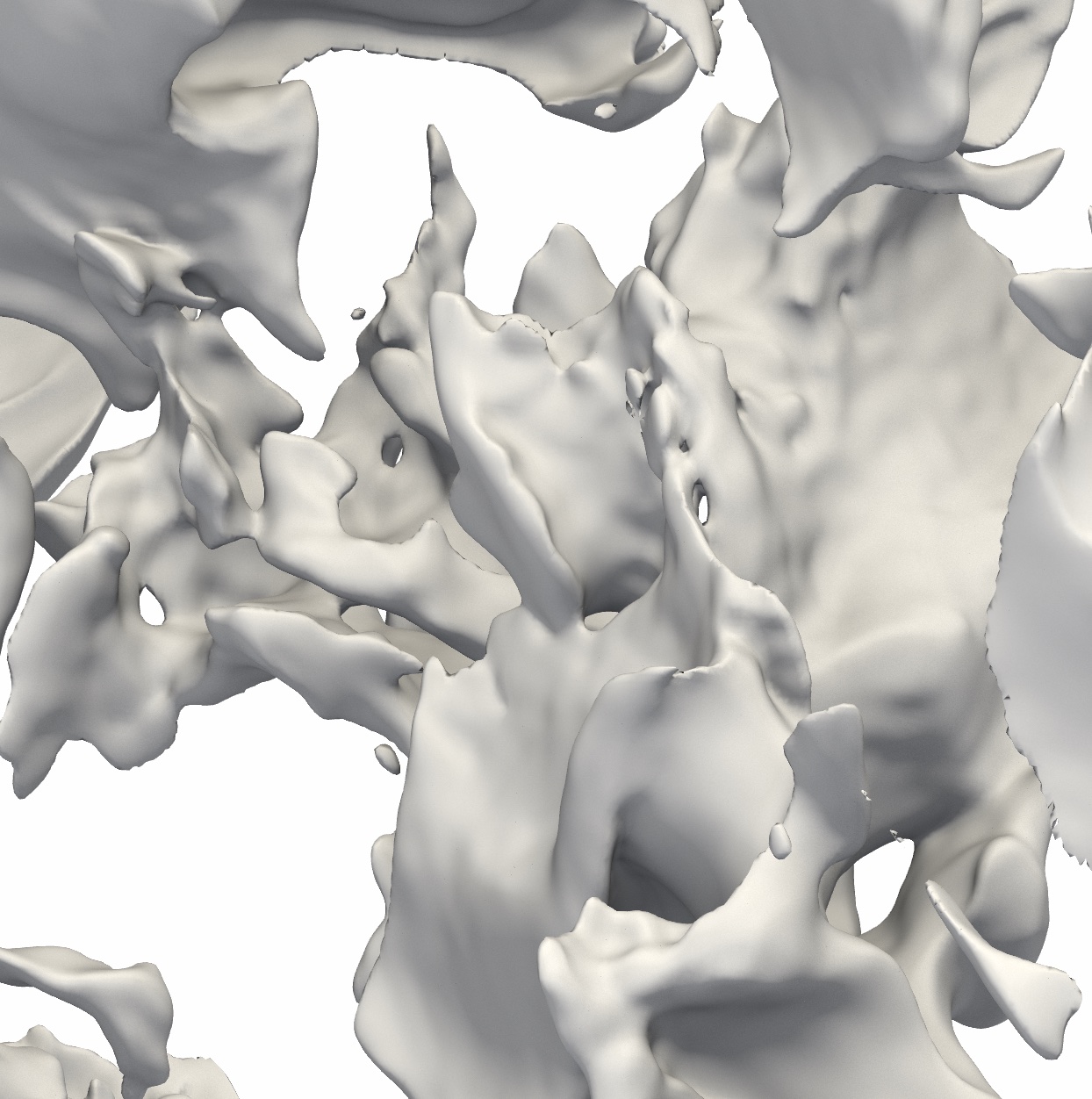}
	}
    \subfloat[\label{subfig:jet_iso_grad}gradient reg.]{
         \includegraphics[width=0.31\linewidth]{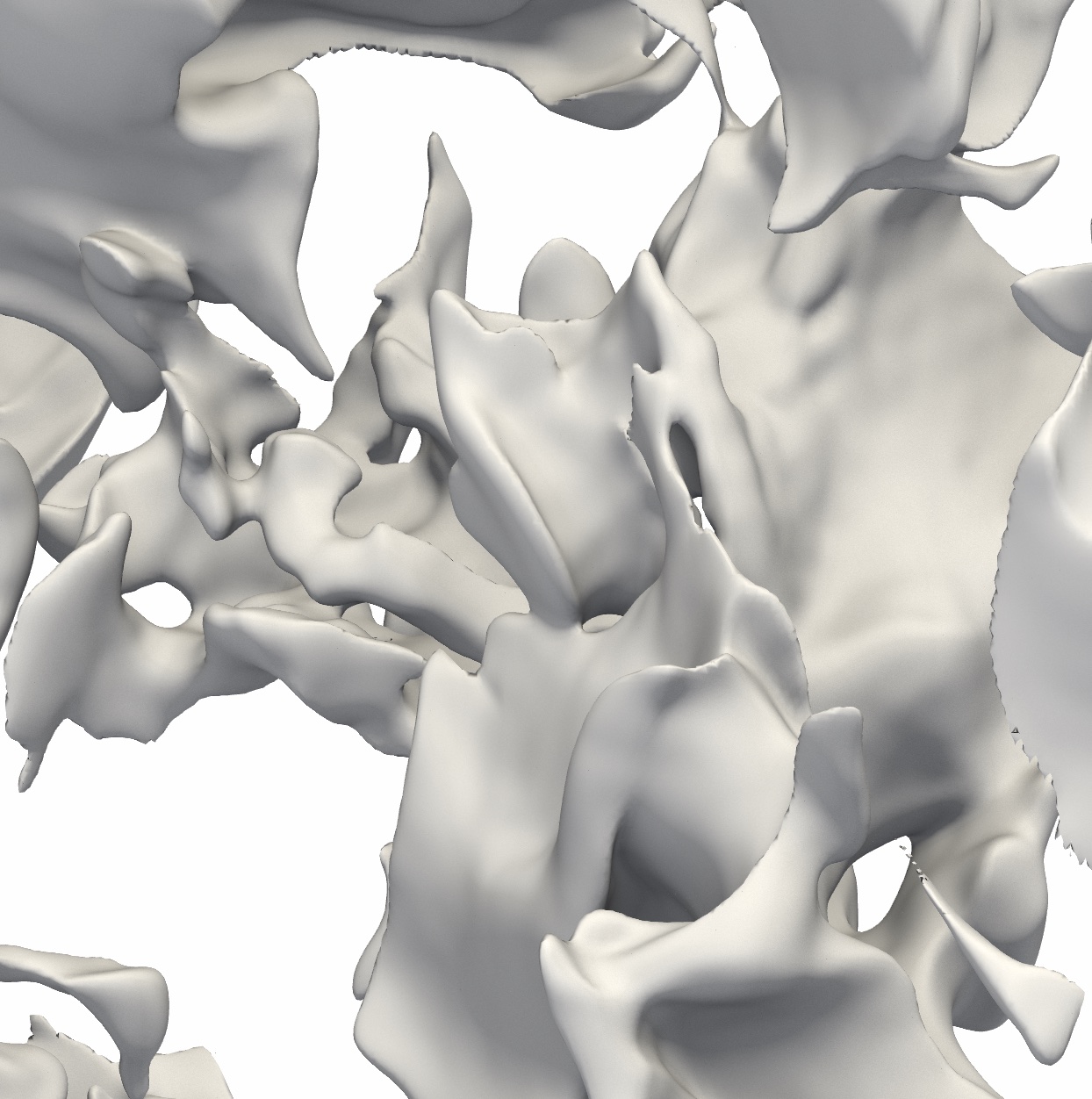}
	}
    \caption{We show isosurfaces for the reconstructed \textsf{jet} volume, comparing with, and without, gradient regularization (c.f. Table~\ref{tab:grad}) in the loss. Note that gradient regularization helps to suppress erroneous high-frequency details.}
    \label{fig:gradreg}
\end{figure}

To show the benefit of gradient regularization we train 2 networks -- one with gradient regularization and one without -- to compress the \textsf{jet} volume, under a compression ratio of $677:1$ for each. Namely, each network is trained with number of weights $m = \frac{C}{200}$, along with weight quantization. We use central differencing to numerically estimate gradients from the provided field. We compare the compression results in terms of the PSNR of the original field, as well as PSNR of the gradient field, following Han et al.~\cite{han2019flow}. The numerically estimated gradient from the provided scalar field is taken as the target gradient field, while we compare to 2 forms of gradients from the decompressed volumes: central differencing of the reconstructed volume (denoted FD-Grad), as well as the analytical gradients obtained by differentiating the network (denoted Net-Grad).

Table~\ref{tab:grad} summarizes the results. We find that incorporating gradient regularization leads to a small decrease in PSNR of the scalar field, but leads to a boost in performance in capturing the target gradient. Interestingly, without gradient regularization, analytical gradients from the network (Net-Grad PSNR) suffer quite a bit relative to numerically-estimated gradients (FD-Grad PSNR). From a visualization perspective, inaccurate gradients can have a large impact on isosurfaces of the scalar field, whose normals point in the same direction as the gradients. We highlight this phenomenon in Fig.~\ref{fig:gradreg}. Note that without gradient regularization, we obtain surfaces with erroneous high-frequency details, whereas gradient regularization leads to smoother surfaces, reflective of ground truth.

\begin{figure}[!t]
    \centering
    \subfloat[\label{subfig:nr_vanilla}Direct Neural Rendering]{
         \includegraphics[width=0.45\linewidth]{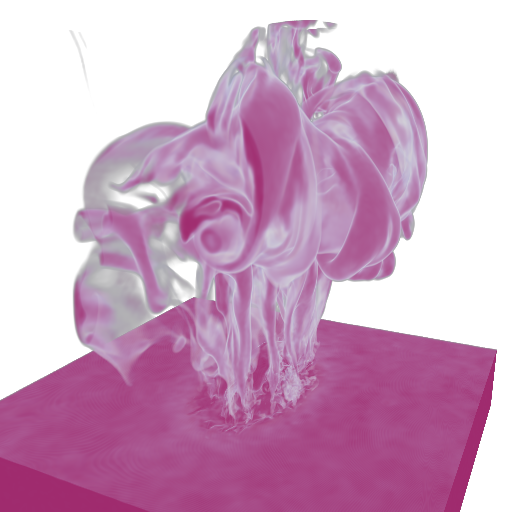}
	}
    \subfloat[\label{subfig:nr_illumination}Gradient-based Shading]{
         \includegraphics[width=0.45\linewidth]{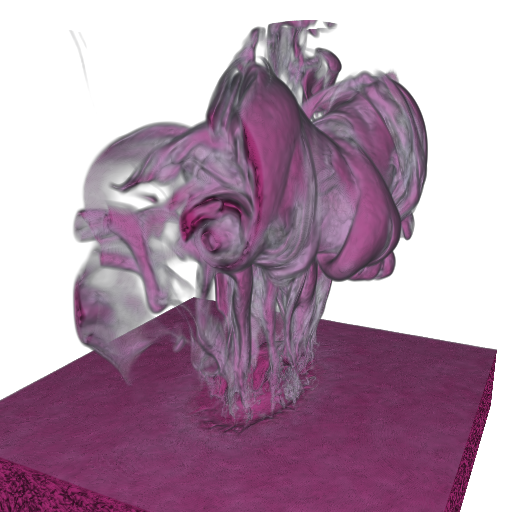}
	}
    \caption{Our neural network can be directly used for volume rendering, rather than reconstructing the volume, shown here for \textsf{asteroid} for (a) a basic volume renderer where our network is evaluated on-demand during ray marching, and (b) the computation of network gradients to enable direct illumination.}
    \label{fig:neuralrender}
\end{figure}

\subsection{Neural Decoder}

Once the network has been trained, and the volume encoded, decoding is rather straightforward, as it amounts to evaluating the neural network. To reconstruct the scalar field we simply evaluate our network at all grid vertices within the volume. This evaluation is relatively efficient given the data parallelism of the network, e.g. a sequence of matrix multiplications and $\sin$ function evaluations.

Beyond reconstruction, a key advantage of neural scalar field representations is the support for random access function evaluation. We may evaluate our function at \emph{any} position, not just grid vertices, hence the network serves as an interpolant. Furthermore, we can compute higher-order information from the network directly, eliminating the need to numerically estimate derivatives. These features allow us to use the network \emph{directly} for visualization purposes, rather than reconstructing the full volume a priori. As an illustration, we have developed a neural volume renderer with our network, where at each step of ray marching, we evaluate our network at all current ray positions to obtain function values, see Fig.~\ref{subfig:nr_vanilla} for an illustration. Moreover, it is trivial to support direct illumination via the computation of function gradients as we ray march, subsequently normalized for shading purposes. We highlight this feature in Fig.~\ref{subfig:nr_illumination}. The quality of the volume-rendered images suggests that our network serves as a good interpolant, not merely overfitting to the sampled field's values.

%-------------------------------------------------------------------------
\section{Experiments}

\begin{figure*}
    \centering
    \includegraphics[width=0.96\linewidth]{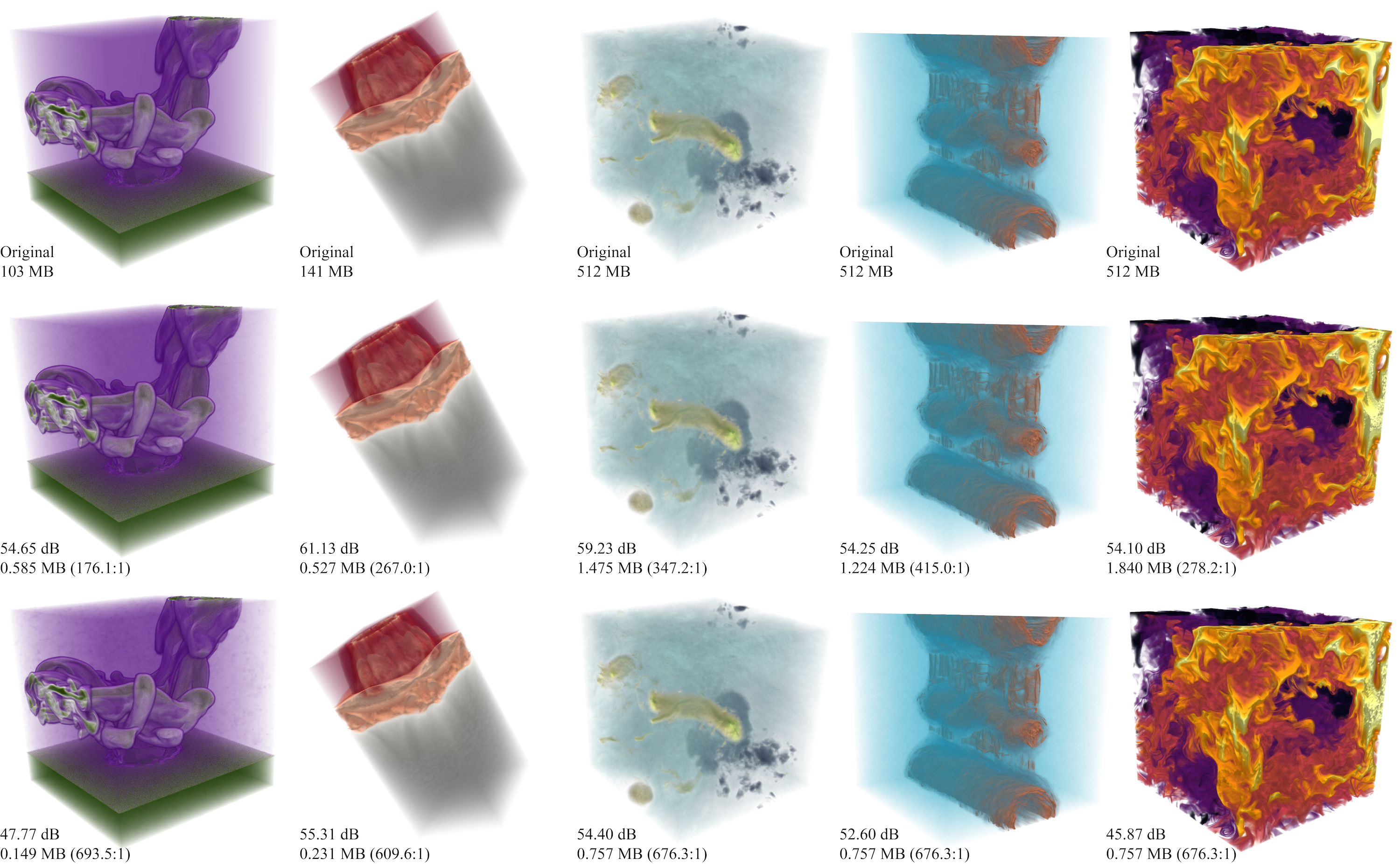}
    \caption{Compression results for our approach, \textsf{neurcomp}.  Left-to-right: \textsf{asteroid} (timestep 0), \textsf{ionization}, \textsf{isotropic\_p}, \textsf{magnetic}, and \textsf{rt}.  Top row: original volumes.  Middle row: medium compression.  Bottom row: high levels of compression.}
    \label{fig:gallery}
\end{figure*}

\begin{figure*}
    \centering
    \includegraphics[width=0.96\linewidth]{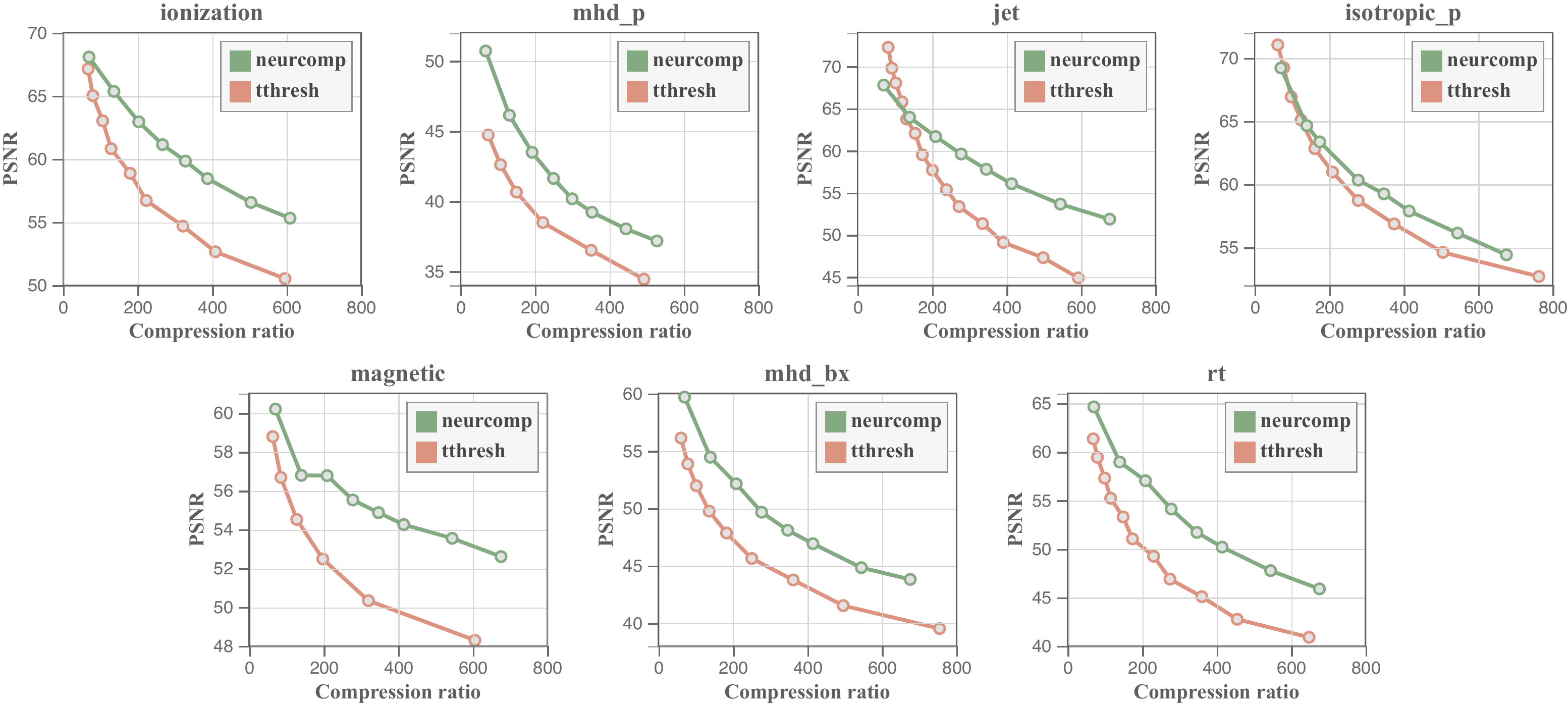}
    \caption{A comparison of our method with \textsf{tthresh}~\cite{ballester2019tthresh} for 3D scalar fields, plotting compression ratio against PSNR.}
    \label{fig:full_results}
\end{figure*}

% forematter: how do we evaluate our approach? datasets?
To evaluate our approach we have run our method on a variety of datasets under different levels of compression; Table~\ref{tab:datasets} summarizes them. Specifically, from the Johns Hopkins Turbulence Database (JHTD)~\cite{li2008public}, \textsf{mhd\_p} and \textsf{mhd\_bx} are respectively pressure and x-coordinate magnetic fields from a magneto-hydrodynamic isotropic turbulence simulation, and \textsf{isotropic\_p} is a pressure field from a forced isotropic turbulence simulation. The \textsf{ionization} volume is a temperature field from an ionization front instability simulation~\cite{whalencompetition}, the \textsf{jet} volume is mixture fraction from a simulation of jet flames~\cite{gyulassy2014stability}, the \textsf{magnetic} volume is from a magnetic reconnection simulation~\cite{magnetic_reconnection}, and \textsf{rt} is from a Rayleigh-Taylor instability simulation~\cite{miranda}. Volumes for JHTD and \textsf{rt} are formed by taking centered spatial crops to access the original field values where the central portion of the simulations take place.

Furthermore, we have evaluated our method on two time-varying volumes \textsf{isabel} and \textsf{asteroid} (Section~\ref{subsec:tv}).  The dataset \textsf{isabel} corresponds to hurricane Isabel's ``QVapor'' field~\cite{wang2004competition} of its first twelve time steps.  Note that we cropped out the ten lowest planes in the $z$-coordinate to remove the NaNs in the data that correspond to land.  The dataset \textsf{asteroid} uses the first ten timesteps from the field ``v02'' from a deep water asteroid impact simulation~\cite{asteroiddeepwater}. 

Fig.~\ref{fig:gallery} shows a gallery of compression results on our approach, which we refer to as \textsf{neurcomp}.  We also compare our method to the state-of-the-art compression technique \textsf{tthresh}~\cite{ballester2019tthresh}, a method for compressing arbitrary tensor data. We limit our comparisons to only \textsf{tthresh} as Ballester et al. have already demonstrated superior results to a wide variety of existing compression methods. Unless otherwise stated, we use OSPRay~\cite{wald2016ospray} for rendering and isosurfacing for visual comparisons.

% Adam optimizer, learning rate varies based on network size, number of iterations -> number of passes over volume, decay learning rate every so many passes
\textbf{Implementation Details.} We use Adam~\cite{kingma2014adam} to optimize our networks, where we found the best results by setting the learning rate to be inversely proportional to the number of weights of the network. Specifically, the learning rate is set as a linear function, varying from $2 \cdot 10^{-5}$ for 5M parameters, to $10^{-4}$ for 800K parameters. We ensure that our network makes a prescribed number of passes over the entire volume, where we found $75$ iterations to be more than enough for convergence, decaying the learning rate by a factor of $5$ every $20$ iterations. At each iteration we set the batch size to fill up the GPU memory, in practice ranging from 16K to 64K grid points from the volume sampled uniformly at random.

\begin{table}[!t]
\centering
\caption{\label{tab:datasets}Name, resolution, precision in bits, and file size in MB for the datasets in our experiments. For time-varying datasets \textsf{asteroid} and \textsf{isabel}, the 4-th dimension of the resolution is the number of timesteps.}
\begin{tabular}{|l|l|l|l|}
\hline
\textbf{Dataset} & \textbf{Resolution}                    & \textbf{Precision} & \textbf{Size} \\ \hline
\textsf{mhd\_p}           & $ 255 \times 255 \times 255$           & float32            & 63            \\ \hline
\textsf{ionization}       & $ 248 \times 248 \times 600$           & float32            & 141           \\ \hline
\textsf{jet}              & $ 512 \times 336 \times 768$           & float32            & 504           \\ \hline
\textsf{mhd\_bx}          & $ 512 \times 512 \times 512$           & float32            & 512           \\ \hline
\textsf{isotropic\_p}     & $ 512 \times 512 \times 512$           & float32            & 512           \\ \hline
\textsf{magnetic}         & $ 512 \times 512 \times 512$           & float32            & 512           \\ \hline
\textsf{rt}               & $ 512 \times 512 \times 512$           & float32            & 512           \\ \hline
\textsf{asteroid}         & $ 300 \times 300 \times 300 \times 10$ & float32            & 1030          \\ \hline
\textsf{isabel}           & $ 500 \times 500 \times 90 \times 12$  & float32            & 1030          \\ \hline
\end{tabular}
\end{table}

\subsection{3D Scalar Fields}
\label{subsec:static}

We first compare our method with \textsf{tthresh} for 3D scalar fields. We run our method over a sequence of compression levels, and plot the resulting compression ratio against PSNR. For these experiments we do not use gradient regularization, but rather, delegate this to Section~\ref{subsec:gradreg}. As \textsf{tthresh} accepts an error/accuracy tolerance (e.g. PSNR) as input parameter, rather than compression ratio, we run \textsf{tthresh} for a sequence of PSNR values such that they give compression ratios that are roughly in the range of what we consider.

Fig.~\ref{fig:full_results} shows the quantitative results. We find that our method is, largely, an improvement over \textsf{tthresh}, with the only exception being the \textsf{jet} volume for low compression ratios. In particular, we find that our method obtains larger gains in performance the higher the compression ratio. This suggests that when our networks are underparameterized (e.g. much fewer parameters than the resolution of the data), they remain robust, and can still obtain good approximations to the given scalar field.

\begin{figure}[!t]
    \centering
    \subfloat[\label{subfig:mhd_bx}Magnetic field ($x$)]{
         \includegraphics[width=0.49\linewidth]{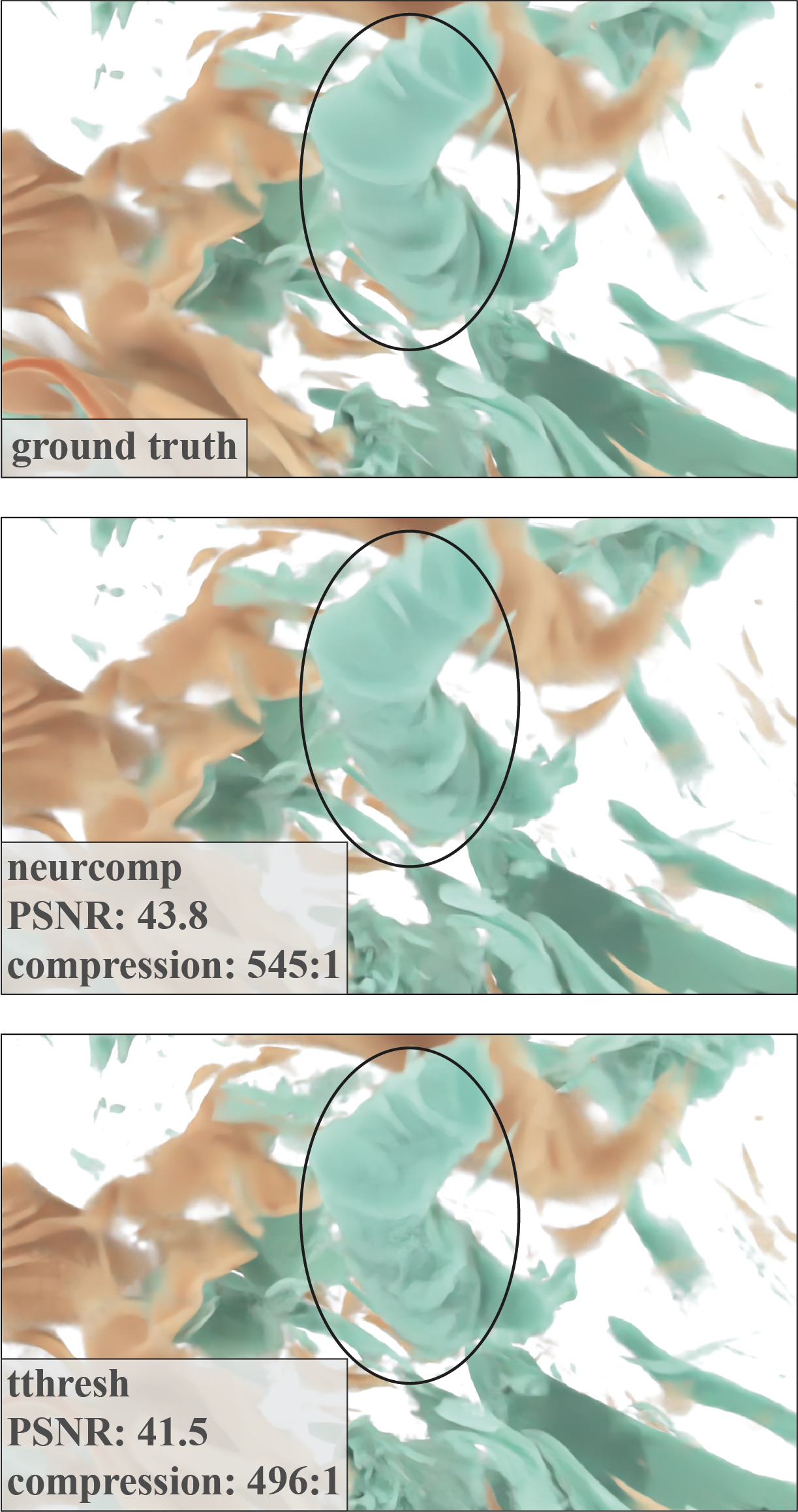}
	}
    \subfloat[\label{subfig:mhd_p}Pressure field]{
         \includegraphics[width=0.485\linewidth]{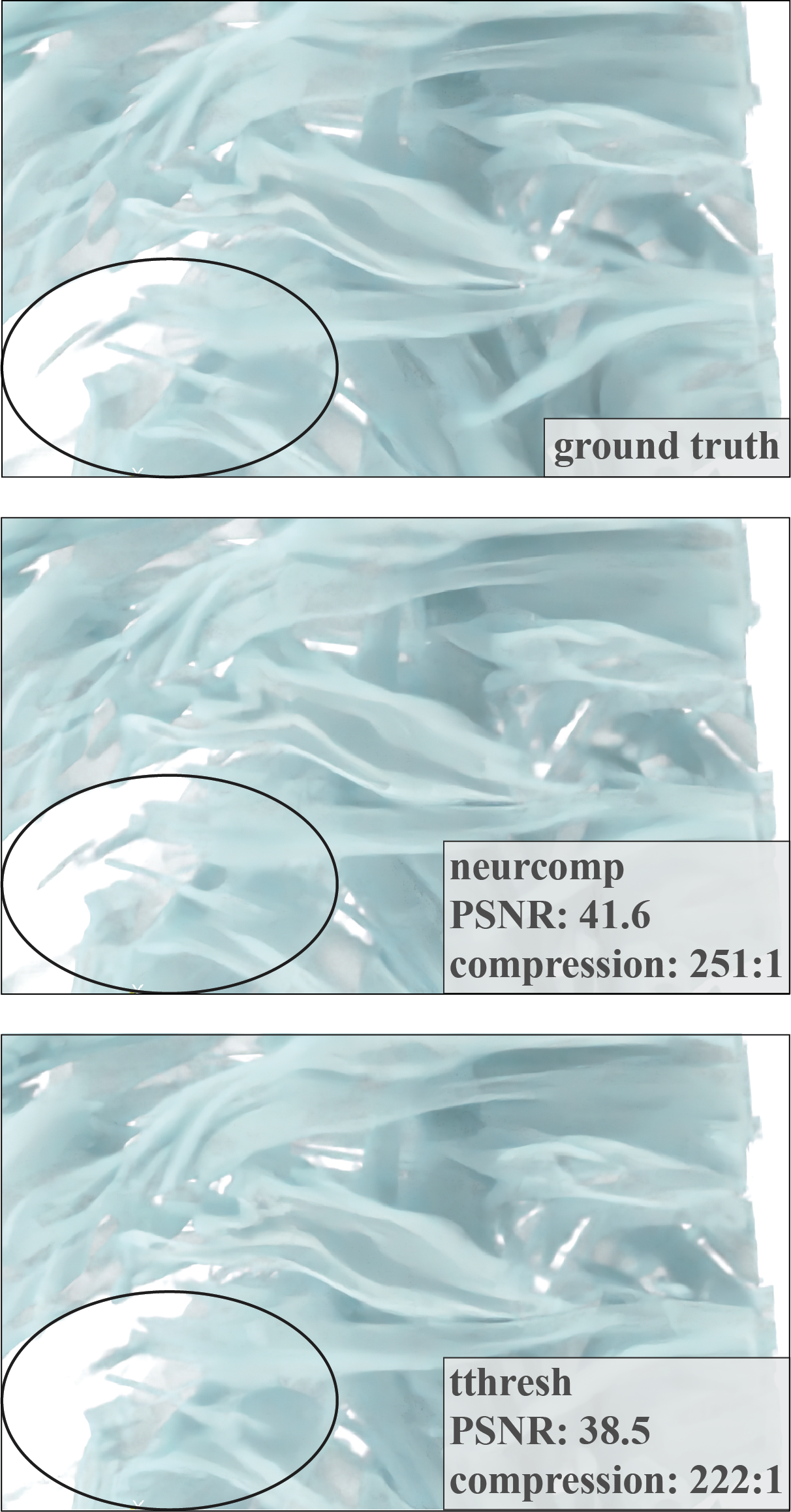}
	}
    \caption{We show visual comparisons between our method and \textsf{tthresh} for the magneto-hydrodynamic simulation. For high compression ratios, we find that our method produces smoother results (a) and can better retain features in the volume (b).}
    \label{fig:mhd_compare}
\end{figure}

\begin{figure}[!t]
    \centering
    \includegraphics[width=0.9\linewidth]{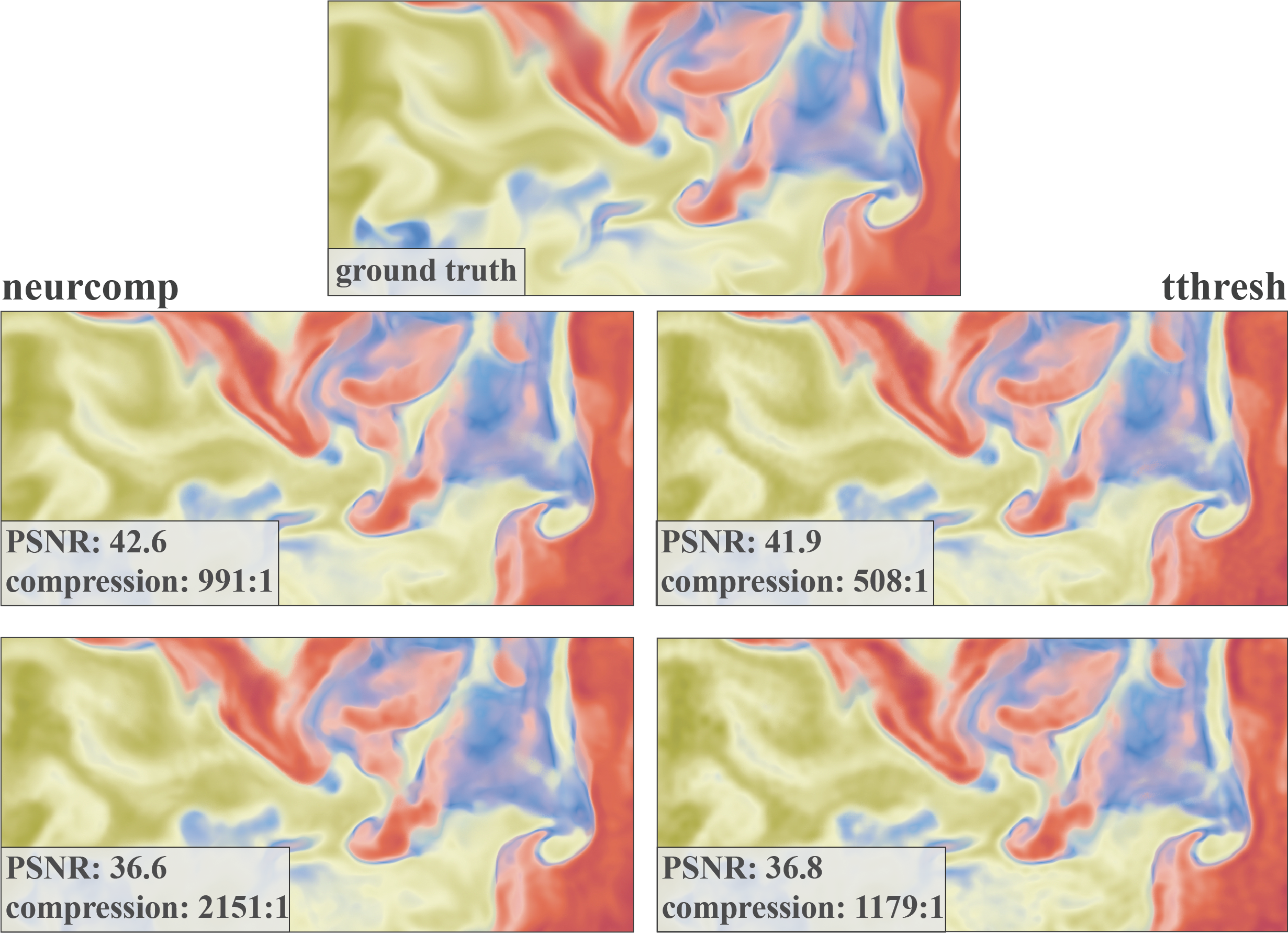}
    \caption{An evaluation of our method under extreme compression ratios for the \textsf{rt} dataset. We find our method gracefully degrades in performance, whereas \textsf{tthresh} tends to reproduce high frequencies.}
    \label{fig:high_comp}
\end{figure}

To further qualitatively assess this robustness in high compression regimes, we visually compare volume renderings of our method with \textsf{tthresh} for the magneto-hydrodynamic simulation, namely its magnetic field (\textsf{mhd\_bx}) and pressure field (\textsf{mhd\_p}) in Fig.~\ref{fig:mhd_compare}. We find that our method is able to reproduce smoother surfaces in the rendering, as shown in the magnetic field plot (a), while in the pressure field (b), we find that components of the volume, corresponding to the chosen transfer function, are better retained through our method. Indeed, we find that our method gracefully degrades for extreme compression ratios, unlike block-based methods~\cite{lindstrom2014fixed,di2016fast} that produce visible artifacts at block boundaries. Global decomposition-based methods~\cite{ballester2016lossy,ballester2019tthresh} also do not suffer from locality issues, but we find that they can produce high-frequency noise in high compression regimes. We demonstrate this in Fig.~\ref{fig:high_comp} for the \textsf{rt} dataset, comparing our method with \textsf{tthresh} under approximately the same PSNR values. We first note that our method produces compressed representations approximately half the size of \textsf{tthresh}. Further, we find that our method produces smoother scalar fields, more reflective of the ground truth as shown on top, unlike the high frequencies reproduced by \textsf{tthresh}.

\begin{figure}[!t]
    \centering
    \includegraphics[width=0.98\linewidth]{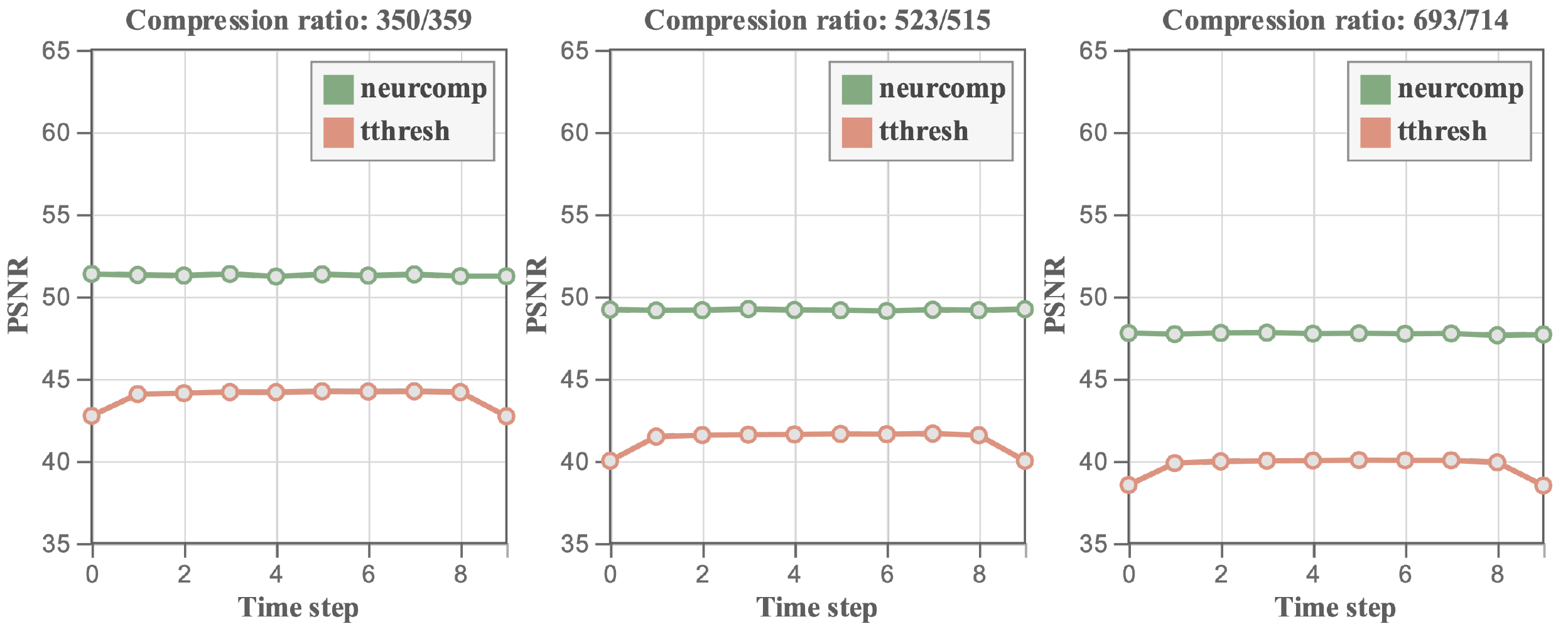}
    \includegraphics[width=0.98\linewidth]{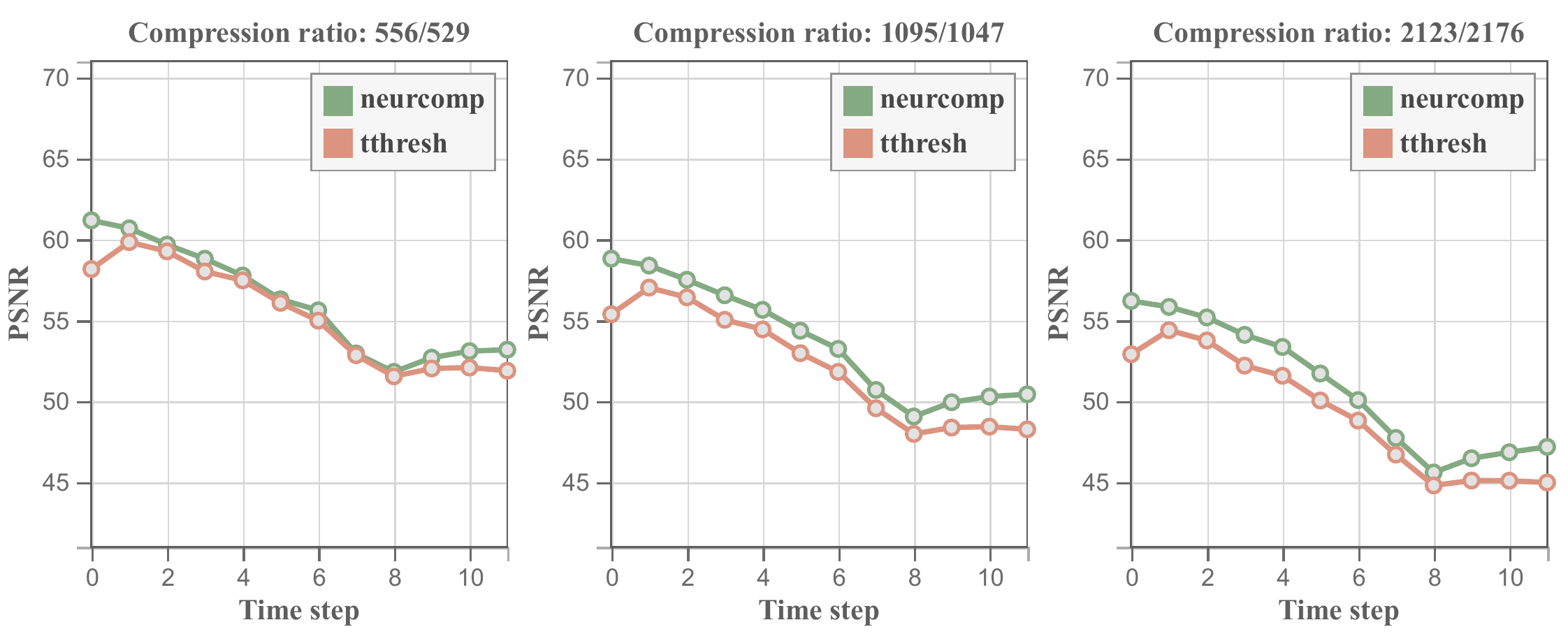}
    \caption{Comparison against \textsf{tthresh} of our approach's performance on the time-varying datasets.  Top row: \textsf{asteroid}, bottom row: \textsf{isabel}.  From left-to-right we vary compression ratios, denoting them as $X/Y$ where $X$ is the compression ratio of \textsf{neurcomp} and $Y$ is the compression ratio of \textsf{tthresh}.\label{fig:tv}}
\end{figure}

\subsection{Time-varying Scalar Fields}
\label{subsec:tv} We evaluate our approach's capability to compress time-varying data by comparing our method with \textsf{tthresh} on \textsf{isabel} and \textsf{asteroid}, assembling the first twelve time steps from \textsf{isabel} and ten consecutive time steps towards the beginning of the \textsf{asteroid} simulation.  While we show results per time-step, we note that both methods treat the data as 4D volumes.
%We used the Deep Water Impact Ensemble Data Set to evaluate our approach's capability to compress time-varying data. 
%Hurricane Isabel dataset is a simulation of a hurricane produced by the Weather Research and Forecast (WRF) model, courtesy of NCAR and the U.S. National Science Foundation (NSF). We use the \textsf{QVAPOR} field for our comparisons, and we took the full $500 \times 500 \times 300$ field over $10$ timesteps.
% kairong: @matt: may need to state which exactly timesteps we used
%The \textsf{asteroid} dataset studies asteroid impacts in deep ocean water. We selected \textsf{v02}, volume fraction water, as the desired scalar field variable for the experiments. 
%We assembled ten consecutive steps of the volume, each of size $300^3$, along a new dimension. The resulting spatial-temporal volume is thus $4$-dimensional with $300^3\times10$ resolution. When running our method on this volume, we sampled $4$-D positions across the whole volume and fed them to the neural nets.
%We assembled $10$ consecutive timesteps of a $300^3$ volume.
% kairong: @matt: may need to state which exactly timesteps we used

\begin{figure}[!t]
    \centering
    \includegraphics[width=0.96\linewidth]{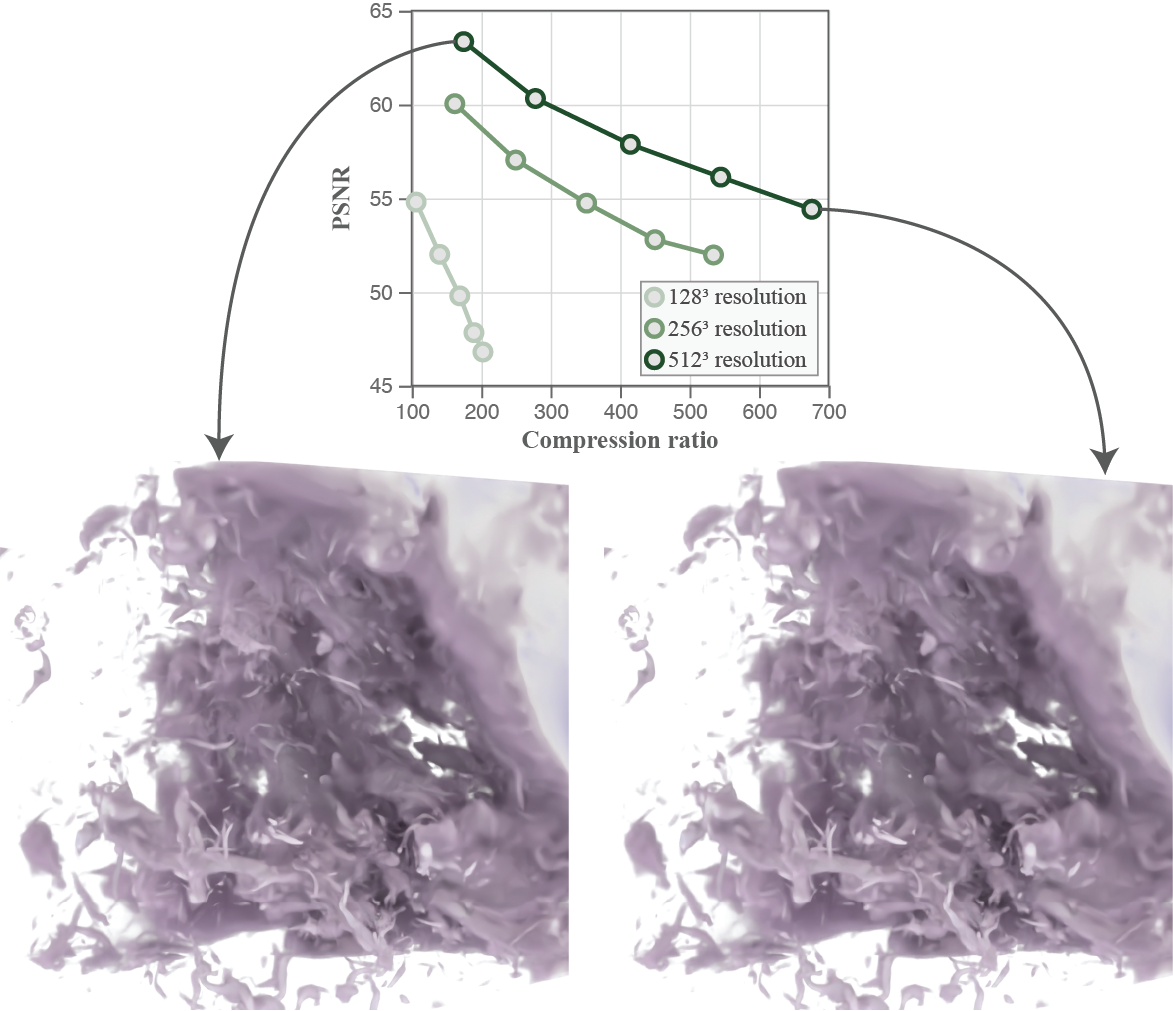}
    \caption{We assess the effect of spatial resolution on performance, evaluating our method on spatial crops of different size for the $512^3$ \textsf{isotropic\_p} volume.}
    \label{fig:resolution}
\end{figure}

Fig.~\ref{fig:tv} show comparisons between our method and \textsf{tthresh} over multiple timesteps with regard to 3 different compression ratios. Note that the compression ratios for our method and \textsf{tthresh} are slightly different, since it is difficult to precisely set the compression ratio for \textsf{tthresh}. As the top of Fig.~\ref{fig:tv} shows, our approach achieves superior results on \textsf{asteroid} dataset when compared with \textsf{tthresh}, with more stability on both ends.  In part, this is due to the fine resolution of sampling in time, where the simulation evolves slowly.
%% TODO UPDATE THIS ONCE TTHRESH RUN FINISHES
%\josh{On the other hand, \textsf{tthresh} performs better with the \textsf{isabel} dataset, as the bottom row of Fig.~\ref{fig:tv} shows (with the exception of the first and last time step). 
% kairong: Need to explain why tthresh performs better with isabel
%However, the difference diminishes as the compression ratio increases, showing preference of our method towards high-compression-ratio scenarios.}
For the \textsf{isabel} dataset, our method performs comparable to \textsf{tthresh} at low compression ratio, except for the start and end timesteps where we outperform, as shown in the bottom of Fig. ~\ref{fig:tv}. For higher compression levels, however, our method gains larger boosts in performance over \textsf{tthresh} at all timesteps, showing the preference of our method for high-compression ratio scenarios.

\subsection{Spatial Resolution}
\label{subsec:res}

Here we examine the impact of spatial resolution on the quality of our compression method. In particular, we wish to gain insight on the following question: do coordinate-based MLPs benefit from high spatial resolution? To test this, we took the \textsf{isotropic\_p} $512^3$ volume, and grabbed centered crops of size $256^3$ and $128^3$. We expect this volume to contain similar statistics for sufficiently large spatial crops, since it corresponds to isotropic turbulence at relatively small scales. We then ran our method on the volumes, setting the network size-to-volume ratio ($\frac{C}{m})$ constant, namely 50, 100, 150, 200, and quantize the network weights to 9 bits. Note that for larger volumes, the storage of the cluster centers leads to overall higher compression ratios, thus for equivalent $\frac{C}{m}$ values lower resolution volumes are given the benefit of the doubt.

Fig.~\ref{fig:resolution} (top) shows the results, plotting performance as PSNR with respect to ground truth. We can see that the higher the spatial resolution, the better the performance, despite (a) the network sizes setup in the same manner, and (b) the higher compression ratios for the larger volumes. We view this as an encouraging property, namely that network capacity is not strictly tied with the spatial resolution. On the bottom of the figure we show volume renderings from low and high-end compression ratios, illustrating that our approach captures the predominant features of the volume even for high levels of compression.

\begin{table}[!t]
\caption{\label{tab:miranda_grad}A comparison of \textsf{tthresh} to our gradient-regularized network.}
\centering
\begin{tabular}{|r|c|c|c|}
\hline
\textbf{Method}    & \textbf{Compression} & \textbf{PSNR} & \textbf{Gradient PSNR} \\ \hline
\textsf{neurcomp} & 545:1 & 47.9         & 50.7          \\ \hline
\textsf{tthresh}   & 258:1 & 47.6         & 46.6          \\ \hline
\end{tabular}
\end{table}

\begin{figure}[!t]
    \centering
    \subfloat[\label{subfig:miranda_iso_gt}ground truth]{
         \includegraphics[width=0.31\linewidth]{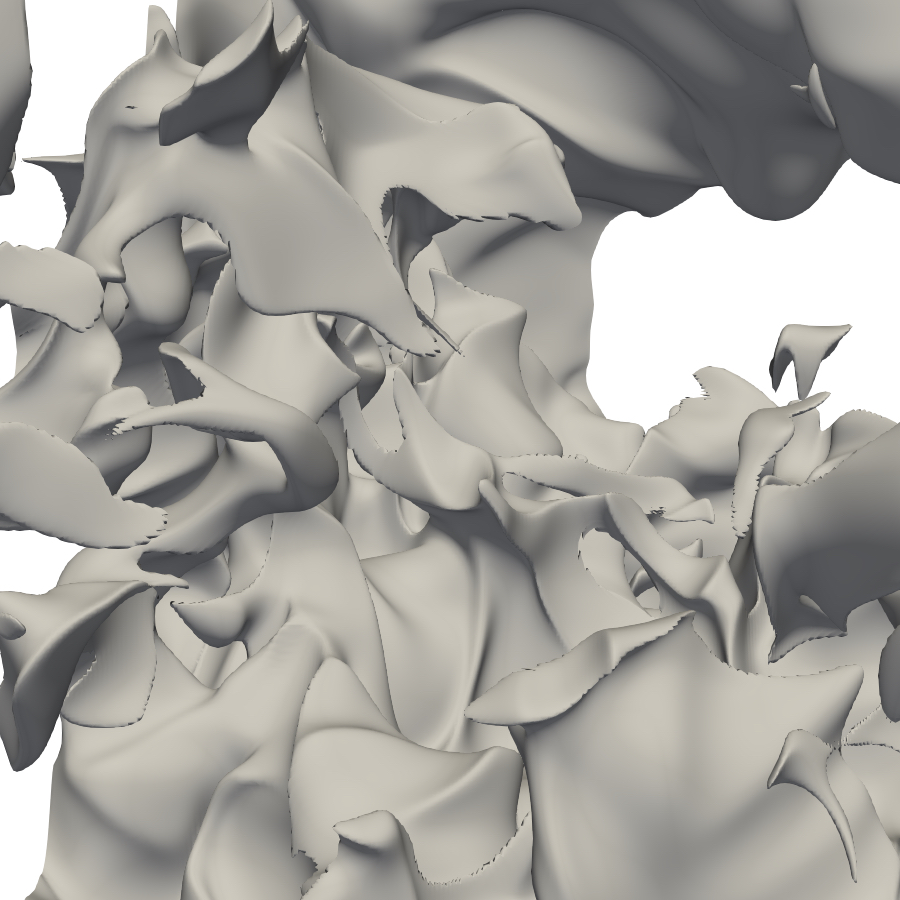}
	}
    \subfloat[\label{subfig:miranda_iso_neurcomp}\textsf{neurcomp}]{
         \includegraphics[width=0.31\linewidth]{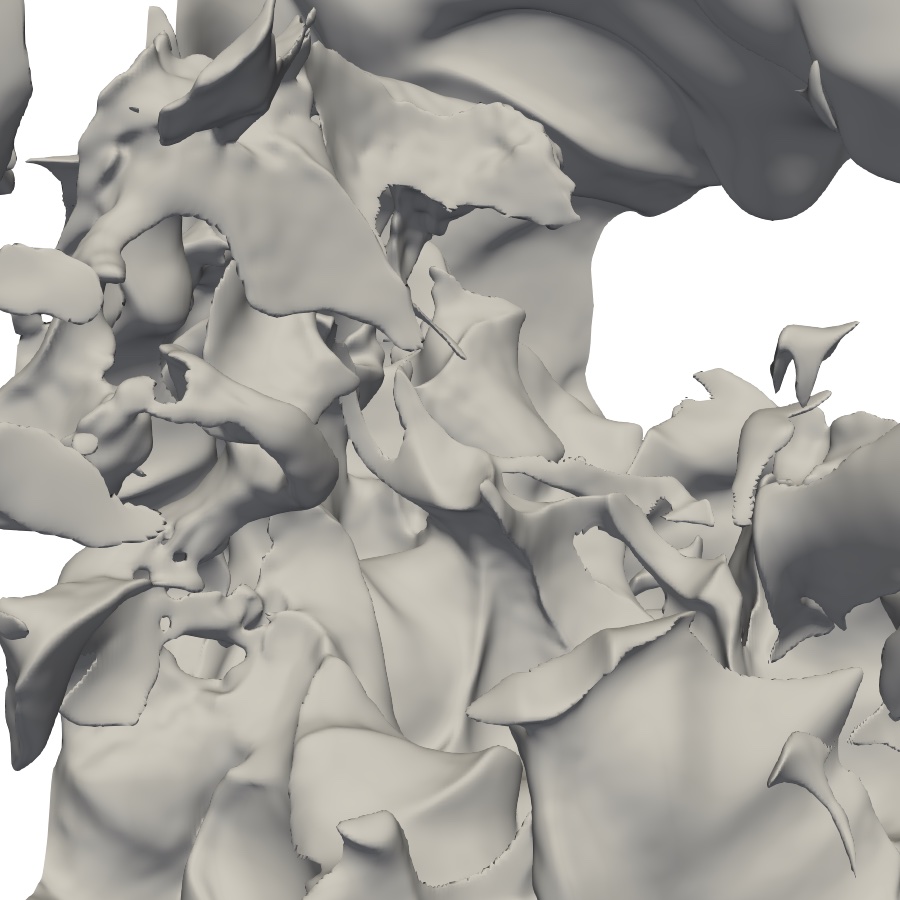}
	}
    \subfloat[\label{subfig:miranda_iso_tthresh}\textsf{tthresh}]{
         \includegraphics[width=0.31\linewidth]{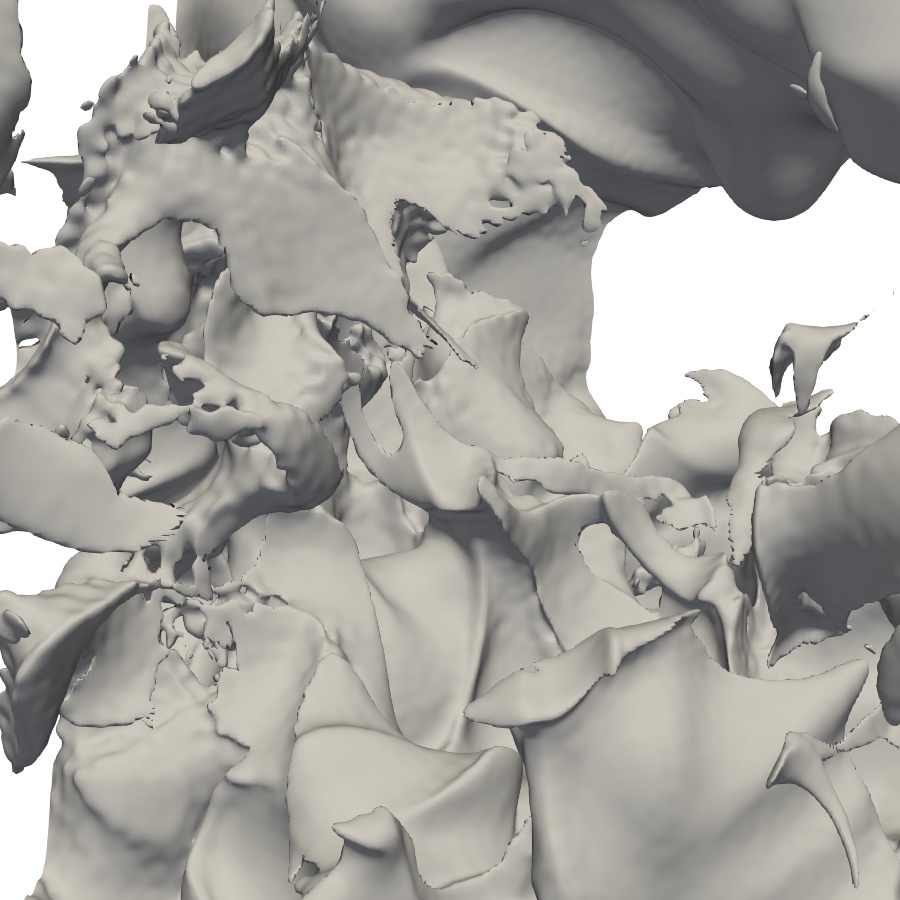}
	}
    \caption{We show isosurfaces for the \textsf{rt} dataset with our gradient-regularized network and \textsf{tthresh}. Though both have roughly the same PSNR, our method suppresses high-frequency details.}
    \label{fig:mirandagrad}
\end{figure}

\subsection{Gradient Regularization}
\label{subsec:gradreg}

Last, we show how gradient regularization can be used to produce better-behaved scalar fields in comparison to prior works. Specifically, for the \textsf{rt} dataset we obtain target gradients through central finite differencing, and train our method using the loss of Eq.~\ref{eq:grad-loss}. We provide our resulting PSNR as a target accuracy for \textsf{tthresh}, yielding comparable PSNR values as shown in Table~\ref{tab:miranda_grad}. However, a gap exists between the methods in terms of the Gradient-based PSNR. As discussed in Sec.~\ref{subsubsec:gradreg}, the effect of gradient regularization will be prominent when isosurfacing the volume, and we can observe in Fig.~\ref{fig:mirandagrad} that \textsf{tthresh} indeed produces isosurfaces that contain considerable noise -- this corroborates findings in Fig.~\ref{fig:mhd_compare}(a) and Fig.~\ref{fig:high_comp}. Indeed, the preservation of higher-order information is nontrivial through existing compression methods, whereas our method can incorporate gradient preservation as part of optimization.

%-------------------------------------------------------------------------
\section{Discussion}

\begin{figure}[!t]
    \centering
    \includegraphics[width=0.53\linewidth]{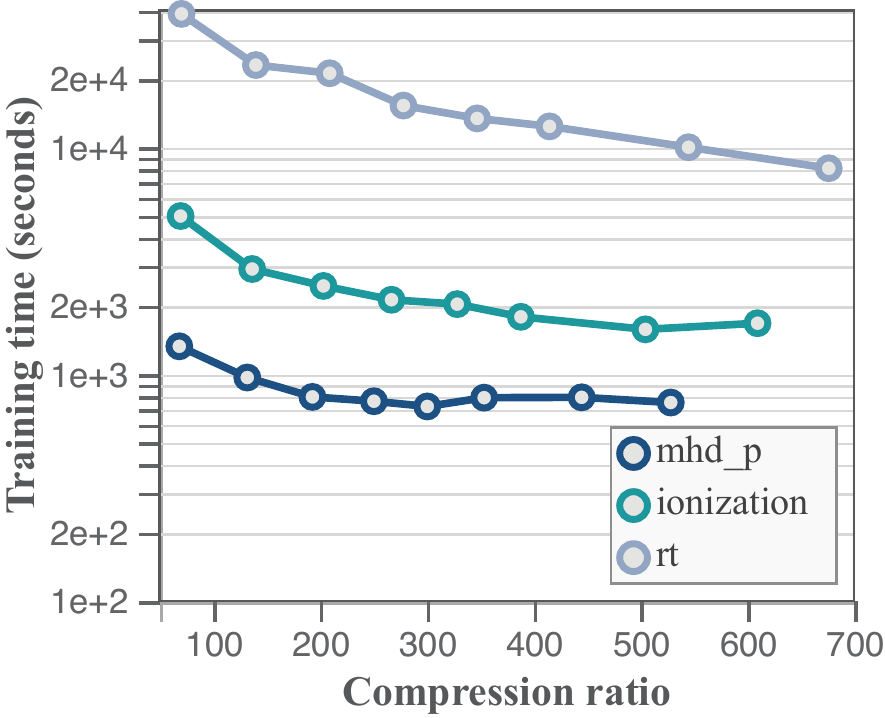}
    \caption{We plot training times for three scalar fields, as a function of compression ratio.}
    \label{fig:timings}
\end{figure}

% computation
Although our method can achieve state-of-the-art results for compressing scalar fields, we acknowledge several limitations. The primary drawback is the high computation time for training. In Fig.~\ref{fig:timings} we plot training times, as a function of compression ratio, for three scalar fields of different spatial resolution -- we observe similar times for different volumes with equivalent resolution.  These times were computed on a single node using an NVIDIA V100 GPU with 16GB of memory.  We foresee our method as being suitable for large simulations running on supercomputers, where as the simulation completes one could run a deep learning job for compression.

We also note that the training times are rather pessimistic -- we find that our network converges to comparable PSNR values in, roughly, $\frac{2}{3}$ the reported times, and so there are opportunities to speed up training. However, as shown, for large volumes training can be quite slow (requiring a few hours per volume).  Interestingly, and perhaps counter-intuitively, as the compression ratio increases it takes less time to compress the data, due to the fact that higher compression ratios utilize smaller network architectures.  So, unlike many standard techniques that start from a lossless perspective and work harder to compress, we control this at the architecture level.

We will address this limitation in future work by investigating multiresolution methods for training~\cite{jiang2020sdfdiff}, adapting our network to take spatial locality into account. A major benefit of our approach to compression is that we do not need to store the entire volume in memory at once, since during training we need only access random samples of the field. Addressing limitations in the time required to train will enable us to realize this benefit.

% generalization
Another limitation related to high computation time is the lack of generalization: the function that is learned is specific to a single volume. We argue that tailoring a neural network to a single volume is precisely what gives us such good performance, as similarly demonstrated in recent work within shape representations~\cite{davies2020overfit}. Nevertheless, we have also shown the efficacy of our method for time-varying scalar fields, and so for future work, we plan to incorporate other factors common to simulation-based data, e.g. data composed of multiple fields and associated with simulation parameters, within our network design. Indeed the notion of generalization is limited for volumetric data within the visualization community, often restricted to fields/time steps produced from a single simulation~\cite{han2019tsr,han2020v2v} or fixing these parameters and instead considering multiple simulation parameters~\cite{he2019insitunet}. Our approach is general enough to accommodate these factors.

\begin{figure}[!t]
    \centering
    \subfloat[\label{subfig:foot_gt}ground truth]{
         \includegraphics[width=0.46\linewidth]{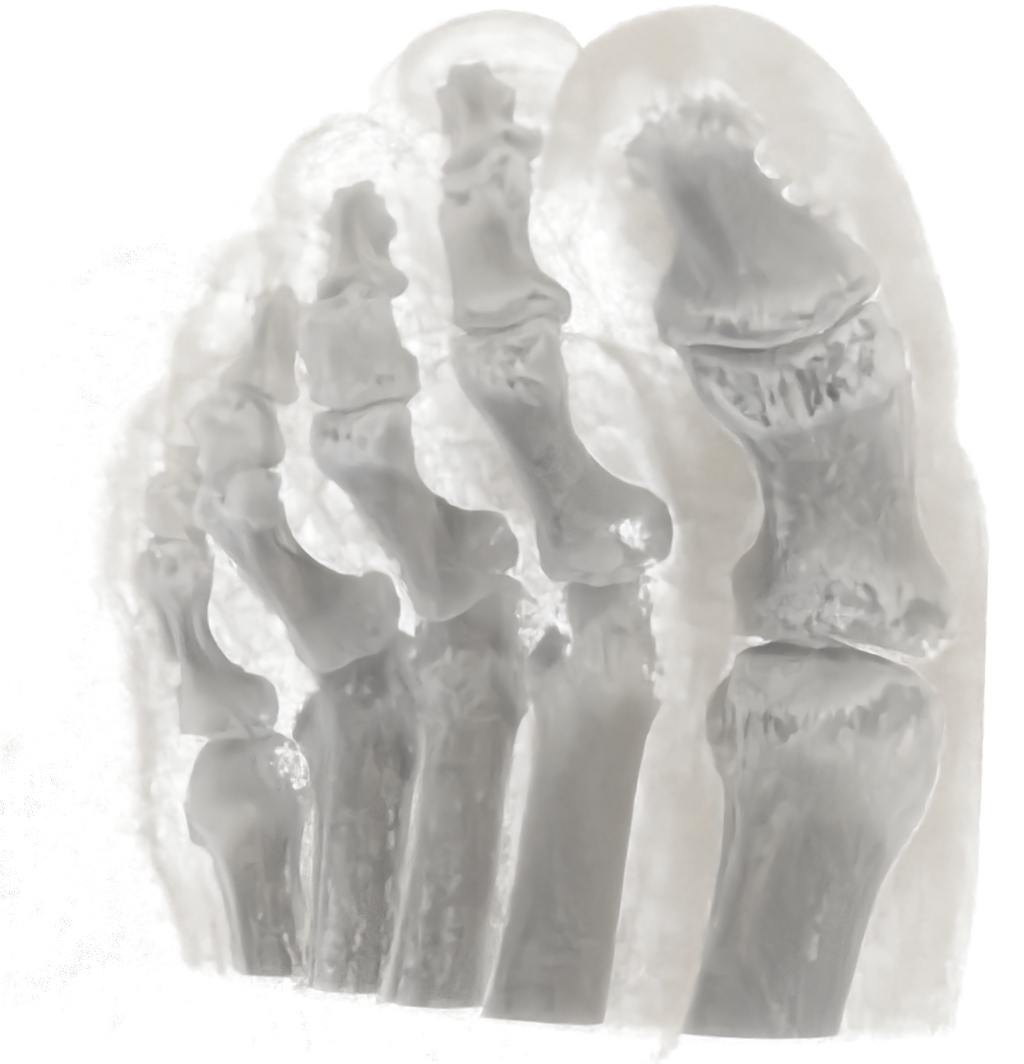}
	}
    \subfloat[\label{subfig:foot_neurcomp}25:1 compression; PSNR: 35]{
         \includegraphics[width=0.46\linewidth]{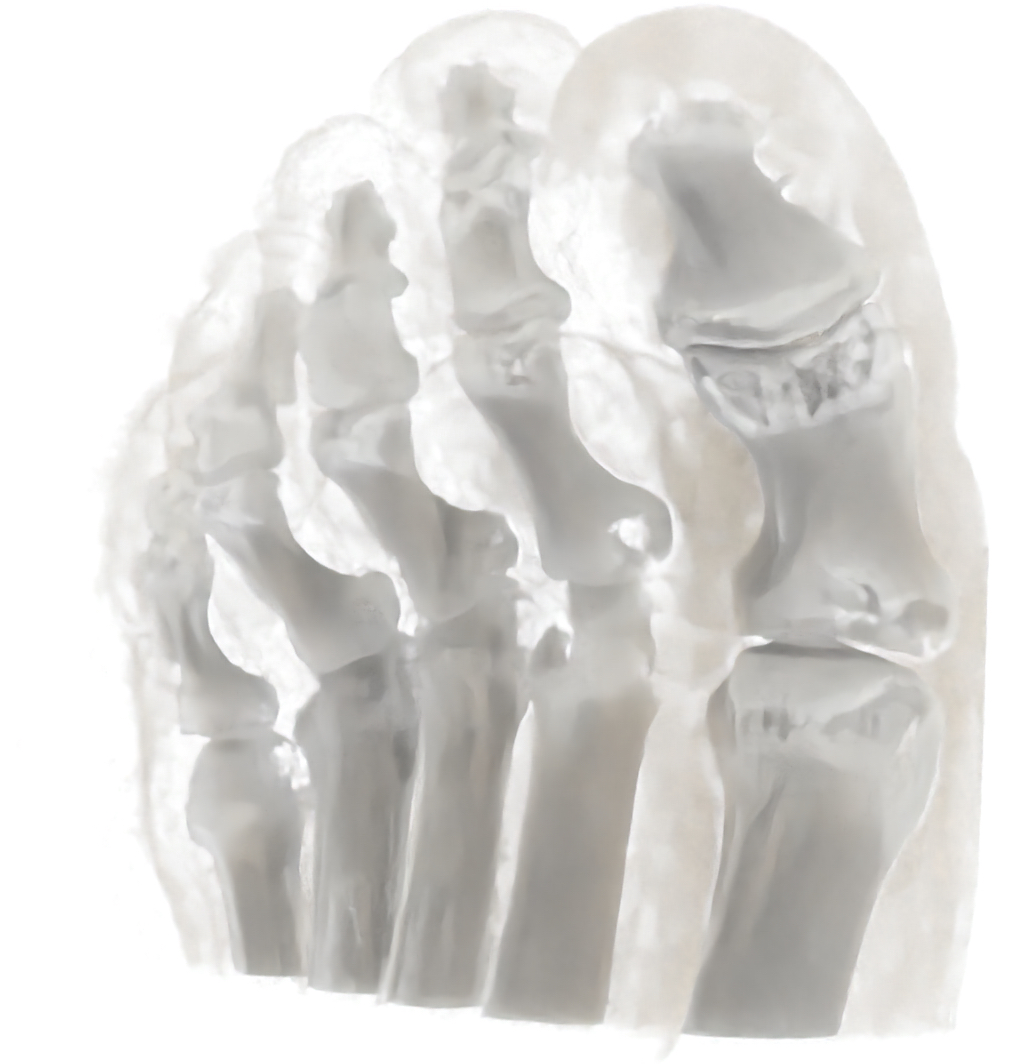}
	}
    \caption{We show our method's performance for a medical image, here in the case of \textsf{foot} ($256^3$, 8-bit precision).}
    \label{fig:foot}
\end{figure}

% other volumes
The use of neural networks for representing volumetric data places some limitations on the types of volumes that we can support. In particular, our method faces limitations in handling noisy volumes, e.g. those produced from medical imaging. The main issue is not in capturing the predominant features of the volume -- our method performs well on this matter, please see Fig.~\ref{fig:foot} for an example on the \textsf{foot} volume. Rather, the use of neural networks makes it challenging to preserve the noise itself. This is also problematic for gradient regularization, where it is challenging to extract useful gradient signal for our network if noise is present. On the other hand, we demonstrated that simple finite difference schemes for simulation data proves helpful for regularization, and we expect higher-order numerical schemes for derivative estimation should also prove useful. Beyond gradients, we can also regularize the learned function with \emph{any} type of differential expression, and this can lead the way to preserving the physics associated with a given simulation as part of our compression approach.

Despite the limitations, we are optimistic about the use of neural networks for scientific data compression. Besides the clear advantages shown over other architectures~\cite{sitzmann2020implicit}, coordinate-based MLPs offer significant flexibility in designing constraints for data preservation.  Future work may show that such networks can better preserve a variety of features in volumetric data. 

%\section{Conclusions}

%-------------------------------------------------------------------------
\section*{Acknowledgements}

This work is supported in part by the National Science Foundation (NSF) under grant numbers IIS-2007444 and IIS-2006710, and by the U.S. Department of Energy, Office of Science, Office of Advanced Scientific Computing Research, under Award Number(s) DE-SC-0019039.

%-------------------------------------------------------------------------
% bibtex
\bibliographystyle{eg-alpha-doi}
\bibliography{eurovis2021}

\end{document}